\documentclass[a4paper,fleqn]{cas-sc}

\usepackage[authoryear,longnamesfirst]{natbib}

\def\tsc#1{\csdef{#1}{\textsc{\lowercase{#1}}\xspace}}
\tsc{WGM}
\tsc{QE}

\usepackage{amsmath,amsfonts}
\usepackage{algorithmic}
\usepackage{algorithm}
\usepackage{array}
\usepackage[caption=false,font=normalsize,labelfont=sf,textfont=sf]{subfig}
\usepackage{textcomp}
\usepackage{stfloats}
\usepackage{url}
\usepackage{verbatim}
\usepackage{graphicx}
\usepackage{cite}

\usepackage{amsmath}
\usepackage{amssymb}
\usepackage{amsfonts}
\usepackage{multirow}
\usepackage{amsthm,color}

\newtheorem{theorem}{Theorem}

\newtheorem{corollary}{Corollary}

\theoremstyle{definition}
\newtheorem{definition}{Definition}

\newtheorem{remark}{Remark}

\newtheorem{assumption}{Assumption}

\newcommand{\change}[1]{#1}

\RequirePackage{amsthm,amsmath,amsfonts,amssymb}

\usepackage{url}            
\usepackage{booktabs}       
\usepackage{amsfonts}       
\usepackage{nicefrac}       
\usepackage{microtype}      
\usepackage{xcolor}         
\usepackage{amsmath,amsthm}
\usepackage{graphicx,psfrag,epsf}
\usepackage{enumerate}

\usepackage{amssymb,color}
\usepackage{algorithm} 
\usepackage{algorithmic} 
\usepackage{multirow}
\usepackage{hyperref}
\usepackage{rotating}
\usepackage{lscape}

\usepackage{threeparttable} 
\usepackage{makecell}
\usepackage{longtable}
\usepackage{rotating}
\usepackage{array}
\usepackage{booktabs}

\def\argmin{\mathop{\rm argmin}}

\def\bfx{\mathop{\bf x}}
\def\bfx{\mathop{\bf X}}

\def\bfx{\boldsymbol{x}}

\begin{document}
\let\WriteBookmarks\relax
\def\floatpagepagefraction{1}
\def\textpagefraction{.001}

\shorttitle{}    

\shortauthors{Wang and Wang}  

\title [mode = title]{Random Projections for Spectral Algorithms: Optimal Rates in Sobolev Norms}  

\tnotemark[1] 

\tnotetext[1]{} 

%




\author[1]{Caixing Wang}[]

\affiliation[1]{organization={School of Statistics and Data Science, Southeast University},
            addressline={}, 
            city={Nanjing},
            postcode={210000}, 
            state={Jiangsu},
            country={China}}

\author[2]{Yue Wang}



\affiliation[2]{organization={Faculty of Business for Science and Technology, School of Management, University of Science and Technology of China},
            addressline={}, 
            city={Hefei},
            postcode={230000}, 
            state={Anhui},
            country={China}}

\cortext[2]{Corresponding author: Yue Wang (wangyue5@ustc.edu.cn)}


\begin{abstract}
Random projections (RP) offer an effective method for reducing computational and storage overhead while preserving the geometric structure of data. However, existing studies have primarily focused on the optimal generalization performance of specific kernel-regularized algorithms with RP in the well-specified setting and under restrictive conditions. In this paper, we provide a comprehensive and improved analysis of the generalization performance of RP-based spectral algorithms under general conditions \change{without sacrificing statistical efficiency}. By leveraging the embedding property of the reproducing kernel Hilbert space (RKHS) and a refined analysis of the operator similarity, we establish optimal learning rates in Sobolev norms that match the minimax lower bounds up to logarithmic factors. For two common implementations of RP, randomized sketches and Nystr\"{o}m sub-sampling (uniform or leverage-based), we show that the projection dimension required for optimality is proportional to the average or maximal effective dimension, yielding a significant reduction in computational cost while maintaining the statistical efficiency.  Our results do not rely on the uniform boundedness assumption on the regression function and are valid across a wide range of source conditions, i.e., $s> \alpha_0-1/\beta$, where $s,\beta$, and $\alpha_0$ denote the smoothness index, capacity index, and the embedding index, respectively. In the benign case when $\alpha_0=1/\beta$, the optimality holds for all $s\in (0,2\tau]$ with $\tau$ and $\gamma$ denoting the qualification index and norm index. Experimental results demonstrate both the theoretical validity and the practical effectiveness of the proposed method. 
\end{abstract}




\maketitle

\section{Introduction}\label{sec:intro}
 In  statistical learning theory, the primary objective is to learn a function that approximates the underlying relationship between input and output variables based on a finite set of training data \citep{vapnik1999overview,vapnik1999nature}. Kernel methods, a class of nonparametric learning algorithms, have gained significant attention due to their ability to capture complex data structures and their strong generalization performance \citep{wahba1990spline,scholkopf2002learning}. These methods operate in a reproducing kernel Hilbert space (RKHS) associated with a positive definite kernel function $K(\cdot,\cdot)$. To address the ill-posedness of such learning problems, regularization techniques are commonly employed, leading to a family of kernel-regularized algorithms, or spectral algorithms \citep{engl1996regularization}. Prominent examples include Tikhonov regularization (also known as kernel ridge regression, KRR) \citep{smale2007learning,caponnetto2007optimal}, spectral cutoff \citep{blanchard2007statistical,dicker2017kernel}, and gradient-based methods \citep{yao2007early,lin2016optimal}. Despite their solid theoretical guarantees, these algorithms often suffer high computational costs when applied to large-scale datasets. For instance, KRR typically requires $O(n^3)$ time and $O(n^2)$ space complexity, where $n$ denotes the number of training samples. To mitigate this issue, various accelerating techniques have been developed, such as distributed algorithms \citep{zhang2015divide,guo2017learning}, random features \citep{rahimi2007random,rudi2017generalization}, randomized sketches \citep{yang2017randomized,zhang2019incremental}, and Nystr\"{o}m sub-sampling \citep{williams2001using,kumar2012sampling}. 

Random projections (RP), including randomized sketches and Nystr\"{o}m sub-sampling as special cases, have emerged as powerful techniques for dimensionality reduction. They offer substantial computational savings while effectively preserving the geometric structure of data \citep{vempala2005random}. The core idea of RP is to estimate the regression function within a suitably chosen random subspace of the RKHS, thereby yielding spectral algorithms with RP. Specifically, for a chosen projection dimension $m \ll n$, we construct a random subspace ${\cal F}=\overline{\text{span}\{K_{\bfx_{i_j}}\}_{j=1}^m}$, where $\{i_1,\ldots,i_m\} \subset \{1,\ldots,n\}$ are randomly selected indices from the input set $\{\bfx_i\}_{i=1}^n$. Performing the spectral algorithm within this subspace ${\cal F}$ corresponds to the Nystr\"{o}m sub-sampling method. More generally, we may consider a random matrix $\mathbf{G} \in \mathbb{R}^{m \times n}$ and define the random subspace as ${\cal F}=\overline{\text{span}\{\sum_{j=1}^n\mathbf{G}_{ij}K_{\bfx_j}, i=1,\ldots,m\}}$, which leads to the randomized sketching method. By choosing $m \ll n$, we can achieve substantial reductions in both time and space complexity while maintaining the statistical efficiency of the original spectral algorithms  \citep{williams2001using,kumar2012sampling,mahoney2011randomized}.

Theoretical analysis of the generalization properties for kernel methods with specific random projection techniques has been extensively studied in the literature. In the fixed design, where the inputs $\{\bfx_i\}_{i=1}^n$ are deterministic, \citet{yang2017randomized} analyzed KRR with randomized sketches and established optimal learning rates in the worst case where the regression function $f_{\rho}$ lies in the RKHS ($s=1$ and $\gamma=0$). Similarly, sharp generalization error bounds in expectation for the Nystr\"{o}m sub-sampling method were derived in \citet{bach2013sharp,alaoui2015fast}. In the random design setting, where the inputs $\{\bfx_i\}_{i=1}^n$ are drawn i.i.d. from an unknown distribution, \citet{rudi2015less,myleiko2019regularized} focused on KRR and general spectral algorithms with Nystr\"{o}m sub-sampling, and established source and capacity dependent learning rates in high probability. However, these results are limited to the well-specified case where $f_{\rho}$ lies in the RKHS ($s\in [1,2\tau]$ with $s$ denoting the smoothness index and $\tau$ denoting the qualification index), and the generalization error is measured via the $L^2_{\rho}$ norm. The most closely related work is \citet{lin2020convergences}, which investigated spectral algorithms with random projections in the mis-specified setting where $f_{\rho}$ may not lie in the RKHS. They derived optimal learning rates in various norms when $s \geq 1-1/\beta$ ($\beta\in(1,\infty)$ with $\beta$ denoting the capacity index), thus covering part of the mis-specified setting ($s\in (0,1)$). However, their analysis relies on restrictive assumptions, such as the H\"{o}lder-type source condition and uniform boundedness of $f_{\rho}$, which may not hold in practice. Until now, it remains an open question whether optimal learning rates of the spectral algorithms with random projections can be achieved  when $s<1-1/\beta$. 

In this paper, we provide a comprehensive and improved analysis of the generalization performance for the spectral algorithms with random projections (SARP) in the mis-specified setting under general conditions \change{without sacrificing statistical efficiency}. The following are the main contributions of this paper:

\noindent \textbf{General conditions.} Our analysis relies on five key assumptions: (i) Bernstein condition on the noise without requiring the uniform boundedness assumption on $f_{\rho}$ (Assumption \ref{ass1}); (ii) capacity condition (Assumptions \ref{ass2} and \ref{ass2_}) and embedding condition (Assumption \ref{ass3}); (iii) general source condition on the regression function $f_{\rho}$ (Assumption \ref{ass4}), of which the H\"{o}lder source condition is a special case. Compared with prior studies on RP-based spectral algorithms \citep{yang2017randomized,rudi2015less,lin2020convergences}, our assumptions are either more general or more readily satisfied in practice.

\begin{figure}
    \centering
    \includegraphics[width=\textwidth]{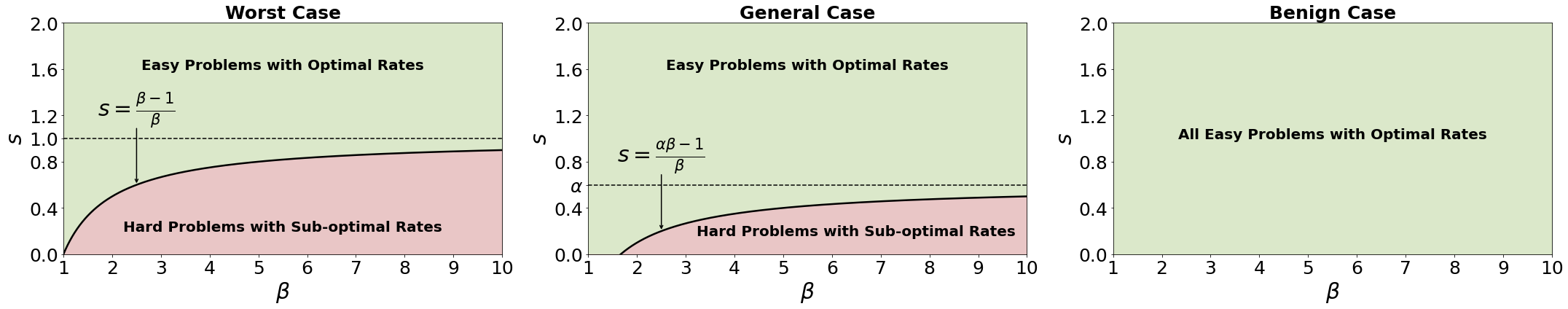}
    \vspace{-5mm}
    \caption{Illustration of the source range for optimality in our work. Panels from left to right represent the worst case ($\alpha=1$), general case ($\alpha\in[{1}/{\beta},1]$), and benign case ($\alpha_0={1}/{\beta}$), respectively.}
    \label{fig:easy_hard}
\end{figure}

\noindent \textbf{Comprehensive and improved learning rates.} Under these more general assumptions, we first establish the minimax lower bounds in Sobolev norms for all $s>0$ in Theorem \ref{low_thm}. For the upper bounds, we combine the integral-operator technique with the embedding property of the RKHS to derive sharp learning rates for general SARP in Theorem \ref{thm1}. By appropriately choosing the regularization parameter $\lambda$, we show that the upper bound is minimax optimal provided the projection error term is of smaller order than the typical approximation and empirical error terms. In Corollary \ref{cor1}, we further prove that this optimality is achievable for both randomized sketches and Nystr\"{o}m sub-sampling, when the projection dimension $m$ is proportional to the average or maximal effective dimension (up to logarithmic factors). \change{Compared to the state-of-the-art results in \citet{lin2020convergences}, our contribution primarily lies in extending the optimality range from $s\geq 1-1/\beta$ to $s> \alpha_0-1/\beta$ for $\alpha_0\in[1/\beta,1]$, rather than improving the convergence exponents already established, as summarized in Figure \ref{fig:easy_hard}. Moreover, our analysis removes the uniform boundedness assumption on $f_{\rho}$ by imposing the Bernstein condition directly on the noise rather than on the response variable. We also extend the analysis from traditional Hölder source conditions to more general source conditions. Regarding the technical analysis, the random projection constructions and their associated concentration bounds largely build on existing studies \citep{rudi2015less,lin2020convergences}. Our new technical contributions lie in the refined operator similarity and empirical error analysis, which enable optimal learning rates under the enlarged source range and weaker boundedness requirements. These improvements also yield smaller sufficient projection dimensions for plain Nyström sampling in certain settings. Further details are provided in Section II of the Supplementary Material.}

\noindent \textbf{Efficient computational improvement.} From a computational perspective, our results show that the projection dimension $m$ required to achieve the optimal learning rates is no larger (and can even be smaller in some cases) than those required in \citet{yang2017randomized,rudi2015less,lin2020convergences}. This leads to a substantial reduction in computational cost while preserving statistical efficiency in a broader source range. The improvement stems from our more refined analysis of operator similarity and sample variance control. 

\change{For a better understanding of our contribution and novelty, we provide a detailed comparison with existing methods in Section II of the Supplementary Material.}

\change{ \textbf{Notations.}
For two positive sequences $\{a_n\}_{n\geq 1}$ and $\{b_n\}_{n\geq 1}$ , we write $a_n \lesssim b_n$ if there exists a constant $C>0$ such that $a_n \leq C b_n$. We write $a_n \asymp b_n$ if both $a_n \lesssim b_n$ and $b_n \lesssim a_n$ hold.
For $a,b\in\mathbb{R}$, let  $a\wedge b=\min\{a,b\}$, $a\vee b=\max\{a,b\}$,
and $a_+=\max\{a,0\}$.   For a bounded linear operator $A$, we denote its operator
norm, adjoint, and range by $\|A\|$, $A^*$, and  $\operatorname{Ran}(A)$, respectively.
For a trace-class operator $A$, $\operatorname{Tr}(A)$
denotes its trace.  The symbol $I$ denotes the identity operator or matrix
on the relevant space.  For $u,v$ in a Hilbert space $\mathcal{H}$, the rank-one
operator $v\otimes u$ is defined by  $(v\otimes u)w=\langle u,w\rangle_{\mathcal{H}}v$
for any $w\in\mathcal{H}$.  The constants $C$ and $c$ may vary from line to line.}

\section{Preliminaries}\label{sec:preliminaries}

Let $({\cal X},{\cal B})$ be a Borel measure space as the input space, and ${\cal Y} \subset \mathbb{R}$ be the output space. Denote by $\rho$ the joint distribution on ${\cal X} \times {\cal Y}$, which can be decomposed as $\rho(\bfx,y)=\rho_{\cal X}(\bfx)\rho(y|\bfx)$ with $\rho(y|\bfx)$ being the conditional distribution of $y$ given $\bfx$.  In
a standard supervised learning framework, we observe a set of training samples $\mathbf{z}=\{(\bfx_i, y_i)\}_{i=1}^n$ drawn i.i.d. from the unknown distribution $\rho$. The objective is to find a function $f$ that minimizes the expected risk over the space of all the measurable functions, which is defined as 
\begin{align}\label{expected_risk}
\mathcal{E}(f)=\int_{\cal X \times \mathcal{Y}}\big(f(\bfx)-y\big)^2d\rho(\bfx,y). 
\end{align} 
The minimizer of the expected risk is the regression function $f_{\rho}(\bfx)=\int_{\cal Y}y d\rho(y|\bfx)$. Consequently, we can define the following nonparametric regression model:
\begin{align}\label{regression_model}
y=f_{\rho}(x)+\varepsilon,
\end{align}
where $\varepsilon$ is the noise satisfying $\mathbb{E}(\varepsilon|\bfx)=0$. 


\subsection{RKHS and Operators} \label{sec:kernel_RKHS}
Consider a bounded and measurable kernel $K:{\cal X} \times {\cal X} \to \mathbb{R}$ that is symmetric and positive definite, i.e., $\sup_{\bfx \in {\cal X}} K(\bfx,\bfx) \le \kappa^2$, $K(\bfx,\bfx^{\prime})=K(\bfx^{\prime},\bfx)$ for $\bfx,\bfx^{\prime} \in {\cal X}$, and $\sum_{i,j=1}^n c_i c_j K(\bfx_i,\bfx_j)\geq 0$ for any $n\in \mathbb{N}$, $\{c_i\}_{i=1}^n \subset \mathbb{R}$, and $\{\bfx_i\}_{i=1}^n \subset {\cal X}$. 
Without loss of generality, we take $\kappa\geq 1$. This kernel is associated with a unique reproducing kernel Hilbert space (RKHS) ${\cal H}=\cal H_{K}$ equipped with inner product $\langle \cdot, \cdot \rangle_{\cal H}$ and the norm $\|\cdot\|_{\cal H}$. The RKHS is the completion of the space of finite linear combinations of the kernel functions $\{K_{\bfx}=K( \bfx,\cdot): \bfx \in \cal X\}$ with respect to the inner product $\langle K_{\bfx}, K_{{\bfx}^{\prime}} \rangle_{\cal H}=K(\bfx,{\bfx}^{\prime})$. For any $f \in {\cal H}_K$, the reproducing property holds:
$$
f(\bfx)=\langle f, K_{\bfx} \rangle_{\cal H}, \quad \forall \bfx \in {\cal X}.
$$ 
Since $K$ is bounded,
$K\in L^2(\rho_X\otimes\rho_X)$.
Consequently, the integral operator $L_K$
is compact, positive, and self-adjoint,
and admits a spectral decomposition in
$L^2_{\rho_X}$. We do not require
$\mathcal X$ to be compact or $K$ to
satisfy the classical Mercer conditions. 

Let $L_{\rho_{\cal X}}^2$ be the space of square-integrable functions from $\cal X$ to $\mathbb{R}$ with respect to $\rho_{\cal X}$, with its norm given by $\|f\|_{\rho}=\sqrt{\int_{\cal X}f^2(\bfx)d\rho_{\cal X}(\bfx)}$. Define the inclusion operator $S_K: {\cal H}_K \rightarrow L_{\rho_{\cal X}}^2$ as $S_Kf(\bfx)=\langle f, K_{\bfx}\rangle_{\cal H}$, for any $f \in {\cal H}_K$ and $\bfx \in \cal X$. We further assume that $S_K$ is injective, i.e., $\ker S_K=\{0\}$. Throughout the paper, we identify $\mathcal H_K$ with $S_K(\mathcal H_K)$ in
$L^2_{\rho_{\mathcal X}}$. Its adjoint operator is denoted by $S_K^*: L_{\rho_{\cal X}}^2 \rightarrow {\cal H}_K$, $
S_K^*g = \int_{\cal X}K_{\bfx}g(\bfx)d\rho_{\cal X}(\bfx),$ for any $g \in L_{\rho_{\cal X}}^2$ and $\bfx \in \cal X$. Then $S_K$ and $S_K^*$ are Hilbert-Schmidt operators and thus compact. Furthermore, we can define the integral operator $L_K$ and the covariance operator $C_K$ as follows:
\begin{equation}\label{integral}
\begin{aligned}
    &L_K: L_{\rho_{\cal X}}^2 \rightarrow L_{\rho_{\cal X}}^2, \quad L_Kg=\int_{\cal X}K(\bfx,\cdot)g(\bfx)d\rho_{\cal X}(\bfx),\\
    &C_K: {\cal H}_K \rightarrow {\cal H}_K, \quad C_K =\int_{\cal X}\langle \cdot, K_{\bfx}\rangle_{\cal H}K_{\bfx}d\rho_{\cal X}(\bfx) =\int_{\cal X}K_{\bfx}\otimes K_{\bfx}d\rho_{\cal X}(\bfx).
\end{aligned}
\end{equation}
Since $K$ is bounded on $\cal X$, the operators $L_K$ and $C_K$ are positive, self-adjoint, and trace-class (hence Hilbert–Schmidt and compact) \citep{steinwart2008support}. They satisfy the identities
$L_K = S_K S_K^*$ and $C_K = S_K^* S_K$. Moreover, as $\sup_{\bfx \in {\cal X}} K(\bfx,\bfx) \le \kappa^2$, we have
$\|S_K\| = \|S_K^*\| \le \kappa$ and $\|L_K\| = \|C_K\| \le \kappa^2$. For any $f \in {\cal H}_K$, the following isometry property holds \citep{steinwart2008support}:
\begin{align}\label{is_pro}
\|f\|_{\rho} = \|S_K f\|_{\rho} = \|C_K^{1/2} f\|_{\cal H}.
\end{align}
In addition, by the polar decomposition of $S_K$, $\|L_K^{-1/2} S_K f\|_{\rho} \le \|f\|_{\cal H}$.

By the spectral theorem, there exists a sequence of non-negative and non-increasing eigenvalues $\{\mu_i\}_{i\geq 1}$, orthonormal basis $\{e_i\}_{i\geq 1}$ of $\overline{\operatorname{Ran}(S_K)} \subseteq  L^2_{\rho_{\cal X}}$, and orthonormal basis $\{\mu_i^{1/2}e_i\}_{i\geq 1}$ of ${\cal H}_K$ such that
\begin{align}\label{op_decomposition}
L_K =\sum_{i=1}^{\infty} \mu_i \langle \cdot, e_i \rangle_{\rho} e_i,   C_K =\sum_{i=1}^{\infty} \mu_i \langle \cdot, \mu_i^{1/2} e_i \rangle_{\cal H} \mu_i^{1/2} e_i.
\end{align}
Moreover, the kernel admits the spectral
representation $K(x,x')=\sum_{i=1}^{\infty}
\mu_i e_i(x)e_i(x')$ with convergence in
$L^2(\rho_X\otimes\rho_X)$. Note that the eigenvalues $\{\mu_i\}_{i\geq 1}$ and eigenfunctions $\{e_i\}_{i\geq 1}$ depend on the kernel $K$ and the marginal distribution $\rho_{\cal X}$.

\subsection{Interpolation Space of RKHS}\label{sec:interpolation_space}
In this section, we introduce the interpolation space (or power space) of the RKHS ${\cal H}_K$. For any $t\geq 0$, we define the fractional integral operator $L_K^{t}$ as $L_K^{t}=\sum_{i=1}^{\infty} \mu_i^{t} \langle \cdot, e_i \rangle_{\rho} e_i$.
\begin{definition}\label{def_interpolation_space}
 Given the RKHS ${\cal H}_K$ associated with the kernel $K$ and the integral operator $L_K$, for any $t\geq 0$, we define the interpolation space $ [{\cal H}_K]_{\rho}^t$ as
 \begin{align}\label{interpolation_space}
 [{\cal H}_K]_{\rho}^t:= \text{Ran} L_K^{t/2}=\left\{\sum_{i=1}^{\infty}a_i\mu_i^{t/2}e_i: \sum_{i=1}^{\infty} a_i^2 < \infty \right\} 
 &\subset L^2_{\rho_{\cal X}}.
 \end{align}
Let $f,g \in  [{\cal H}_K]_{\rho}^t$ and $f=\sum_{i=1}^{\infty}a_i\mu_i^{t/2}e_i, g=\sum_{i=1}^{\infty}b_i\mu_i^{t/2}e_i$, define the inner product and norm of $ [{\cal H}_K]_{\rho}^t$ as $\langle f,g \rangle_{ [{\cal H}]_{\rho}^t}=\langle L_K^{-t/2}f, L_K^{-t/2}g \rangle_{\rho}=\sum_{i=1}^{\infty} a_i b_i,   \|f\|_{ [{\cal H}]_{\rho}^t}=\langle f,f \rangle_{ [{\cal H}]_{\rho}^t}^{1/2}=(\sum_{i=1}^{\infty} a_i^2)^{1/2}$.
\end{definition}

The interpolation space $[{\cal H}_K]_{\rho}^t$ is also a separable Hilbert space with  orthonormal basis $\{\mu_i^{t/2}e_i\}_{i\geq 1}$. It follows immediately that $ [{\cal H}_K]_{\rho}^1= {\cal H}_K$ and $ [{\cal H}_K]_{\rho}^0 \subseteq L^2_{\rho_{\cal X}}$. For any $0 \leq t_1 \leq t_2$, there is a continuous embedding $[{\cal H}_K]_{\rho}^{t_2} \hookrightarrow [{\cal H}_K]_{\rho}^{t_1} \hookrightarrow [{\cal H}_K]_{\rho}^{0}$ \citep{fischer2020sobolev}. The interpolation space $[{\cal H}_K]_{\rho}^s$ provides a natural framework for describing the regularity of the regression function $f_{\rho}$. For example, if $f_{\rho} \in [{\cal H}_K]_{\rho}^t$ for some $t>1$, then $f_{\rho}$ is smoother than functions in the RKHS ${\cal H_K}$; Conversely, when $0<t<1$, the condition
$f_\rho\in[{\cal H}_K]_\rho^t$ is weaker and may allow $f_\rho\notin{\cal H}_K$.

For $0<t<1$, the interpolation space $[{\cal H}_K]_{\rho}^t$ can be characterized by the real interpolation method \citep{tsybakov2008nonparametric,steinwart2012mercer,bitzer2025spectral}. Specifically, we have
\begin{align}\label{real_interpolation}
[{\cal H}_K]_{\rho}^t \cong [L^2_{\rho_{\cal X}}, [{\cal H}_K]_{\rho}^1]_{t,2},
\end{align}
where $\cong$ denotes the isomorphism between the two spaces and $[\cdot,\cdot]_{t,2}$ is the real interpolation functor of two function spaces.

\subsection{General Source Condition}

In statistical learning theory, it is standard to restrict the function class of the underlying problem. A common approach is to consider a family of probability measures $\mathcal{P}(\Omega)$ whose associated regression function satisfies $f_{\rho} \in \Omega$, where $\Omega$ is a hypothesis set. Typically, $\Omega$ is compact and specified by regularity (smoothness) conditions. In the context of kernel methods, a widely used regularity condition is the Hölder source condition \citep{caponnetto2007optimal,steinwart2009optimal,guo2017learning} in terms of the integral operator $L_K$:
\begin{align}\label{holder}
     f_{\rho}\in \Omega_{s, R}:=\left\{f \in L^2_{\rho_{\cal X}}: f=L_K^{s/2}g, \|g\|_{\rho}\leq R \right\},
\end{align}
where $s>0$ is the smoothness parameter. When $s=1$, the source condition means that $f_\rho$ admits a representative in $\mathcal H_K$, with minimum RKHS norm
at most $R$. We denote the function class of $f_{\rho}$ as ${\cal F}=L_K^{s/2}(L^2_{\rho_{\cal X}})$, and note that ${\cal H}_K = L_K^{1/2}(L^2_{\rho_{\cal X}})$, and $L_K^{s_1/2}(L^2_{\rho_{\cal X}}) \subseteq L_K^{s_2/2}(L^2_{\rho_{\cal X}})$ \citep{steinwart2008support} for $s_1 \geq s_2$.  When $s \geq 1$, the functional class $\cal F$ is a subset of the assumed RKHS ${\cal H}_K$, so we have $f_{\rho} \in {\cal H}_K$ (well-specified case). When  $s \in (0,1)$,  the functional class $\cal F$ is larger than the assumed RKHS ${\cal H}_K$, and there exist  some cases where $f_{\rho} \notin {\cal H}_K$ (mis-specified case). By \eqref{interpolation_space}, the source condition in \eqref{holder} is equivalent to $f_\rho\in[\mathcal H]^s_\rho$ and $\|f_\rho\|_{[\mathcal H]^s_\rho}\le R$.

The Hölder source condition characterizes the regularity of $f_{\rho}$ through the decay of the coefficients in the eigenbasis of $L_K$. Specifically, if $\{(\mu_i,e_i)\}_{i\geq 1}$ are the eigenpairs of $L_K$, then $f_{\rho}$ admits the expansion $f_{\rho}=\sum_{i=1}^{\infty}\langle f_{\rho},e_i\rangle_{\rho}\ e_i$. The Hölder source condition implies that the coefficients $\langle f_{\rho},e_i\rangle_{\rho}$ decay at a rate determined by $s$, i.e., $
\sum_{i=1}^{\infty}\frac{\langle f_{\rho},e_i\rangle_{\rho}^2}{\mu_i^{s}}  \leq R^2,$
which is known as Picard's criterion \citep{bauer2007regularization}. 
A natural generalization is to replace the power function $\mu_i^{s}$ by a more flexible index function $\phi(\cdot)$, leading to the more general condition $
\sum_{i=1}^{\infty}\frac{\langle f_{\rho},e_i\rangle_{\rho}^2}{\phi^2(\mu_i)}  \leq R^2.$
Consequently, we define the general source condition as
\begin{align*}
 f_{\rho}\in \Omega_{\phi,s, R}:=\left\{f \in L^2_{\rho_{\cal X}}: f=\phi(L_K)g, \|g\|_{\rho}\leq R, \phi\in C_s \right\},
\end{align*}
where $\mathcal{C}_s$ is the index function class. In this paper, we consider 
\begin{equation}\label{cs_class}
\begin{aligned}
\mathcal{C}_s=&\big\{\phi: [0,\kappa^2]\rightarrow \mathbb{R}^+ \big| \phi \text{ is  continuous and non-decreasing, } \phi(0)=0,     0<\phi(t)<\infty  \ \text{for} \ t\in(0,\kappa^2], \\
& \qquad \qquad \qquad \qquad \text{ and } \change{\frac{ { \phi^2(t)}}{t^{s}} }\text{ is non-decreasing for some } s>0  \big\}.
\end{aligned}
\end{equation}
A typical example is the H\"{o}lder source condition, which corresponds to the choice $\phi(t)=t^{s/2} \in C_s$.

\section{Spectral Algorithms and Random Projections}\label{sec:spectral_RP}

Spectral algorithms (SA) comprise a family of regularization methods that leverage the spectral operator through filter functions to stabilize learning problems \citep{engl1996regularization}. Initially developed for ill-posed linear inverse problems \citep{de2005learning}, they have been subsequently extended to nonparametric regression via the connection between learning theory and inverse problems \citep{rosasco2005spectral,de2006discretization,guo2017learning}.

Suppose that we have a training dataset $\{(\bfx_i,y_i)\}_{i=1}^n$ drawn i.i.d. from an unknown distribution $\rho$. Define the sampling operator 
\begin{align*}
S_{K,n}: {\cal H}_K \rightarrow \mathbb{R}^{n},\quad S_{K,n}f=\frac{1}{\sqrt{n}}\big(\langle f, K_{\bfx_1}\rangle_{\cal H}, \ldots, \langle f, K_{\bfx_{n}}\rangle_{\cal H}\big),
\end{align*}
and its adjoint operator   
\begin{align*}
S_{K,n}^*: \mathbb{R}^{n} \rightarrow {\cal H}_K, \quad S_{K,n}^*\boldsymbol{\alpha} = \frac{1}{\sqrt{n}}\sum_{i=1}^{n}\alpha_i K_{\bfx_i},
\end{align*}
for $\boldsymbol{\alpha}=(\alpha_1, \alpha_2, \ldots, \alpha_n)^{T} \in \mathbb{R}^{n}$.
 The empirical covariance operator is then
\begin{align*}
C_{K,n}: {\cal H}_K \rightarrow {\cal H}_K, \quad C_{K,n} =S_{K,n}^* S_{K,n}=\frac{1}{n}\sum_{i=1}^{n} K_{\bfx_i} \otimes K_{\bfx_i}.
\end{align*}
Our goal is to learn an estimator $\hat{f} \in {\cal H}_K$ from the training data that approximates the true regression function $f_{\rho}$ defined in \eqref{regression_model}, i.e. the empirical
risk minimizer, $\hat{f} = \argmin_{f \in {\cal H}_K} \frac{1}{n}\sum_{i=1}^n (f(\bfx_i)-y_i)^2$.
Using the operator notations introduced above, the minimizer should satisfy the normal equation $C_{K,n}\hat{f}=S_{K,n}^*\mathbf{y}$, where $\mathbf{y}=\frac{1}{\sqrt{n}}(y_1, y_2, \ldots, y_n)^{T}$. However, directly solving the normal equation is prone to overfitting, and the operator $C_{K,n}$ is often singular.

To address these issues, spectral algorithms introduce a regularization mechanism by applying a filter function to the empirical covariance operator. This stabilizes the solution and mitigates overfitting through the regularization effect of the filter function. Specifically, a spectral algorithm produces an estimator  $f_{n,\lambda}$ of the form
\begin{align}\label{spectral_estimator}
	f_{n,\lambda}=g_{\lambda}(C_{K,n})S_{K,n}^*\mathbf{y},
\end{align}
where $\lambda>0$ is a regularization parameter and $g_{\lambda}:[0,\kappa^2] \to \mathbb{R}$ is a filter function satisfying certain conditions (see Definition \ref{def_filter} below). 
\begin{definition}\label{def_filter}
	A function $g_{\lambda}(\cdot): [0,\kappa^2] \rightarrow \mathbb{R}^{+}$ indexed by $\lambda \in \Lambda \subset \mathbb{R}^{+}$ is called a filter function with qualification $\tau\geq 1$, if it satisfies the following conditions: 
	\begin{align}\label{filter_condition1}
	&\sup_{\nu \in [0,1]}\sup_{\lambda \in \Lambda} \sup_{t \in [0,\kappa^2]}|t^{\nu}g_{\lambda}(t)|\lambda^{1-\nu}\leq E,  \quad \sup_{\nu \in [0,\tau]}\sup_{\lambda \in \Lambda} \sup_{t \in [0,\kappa^2]}|(1-tg_{\lambda}(t))|t^{\nu}\lambda^{-\nu}\leq F_{\tau},
	\end{align}
	where $E$ and $F_{\tau}$ are positive constants independent of $\lambda$ and $t$.
\end{definition}

Different choices of the filter function $g_{\lambda}$ yield different spectral regularization schemes. 
In Table~\ref{tab:spectral-filters}, we list several common spectral algorithms together with their corresponding filter functions, qualifications, and the associated constants $(E,\,F_{\tau})$.

\begin{table*}[t]
	\centering
	\caption{Several common spectral algorithms, along with their corresponding filter functions, qualifications, and the associated constants.}
	\label{tab:spectral-filters}
	\begin{threeparttable}
			\begin{tabular}{l l l c c}
				\hline
				Method & Filter Function $g_{\lambda}(t)$ & Parameters & Qualification $\tau$ & $(E,\,F_{\tau})$ \\
				\hline
				\textbf{Tikhonov Regularization (KRR)} &
				$ \dfrac{1}{t+\lambda} $ &
				$\lambda>0$ &
				$1$ &
				$(1,\,1)$ \\
				
				\textbf{Iterative Regularization} &
				$ \displaystyle \sum_{k=1}^{l}\lambda^{k-1}(\lambda+t)^{-k}
				$ &
				$l\in\mathbb{N}$ &
				$l$ &
				$(l,\,1)$ \\
				
				\textbf{Gradient methods (GM)} &
				$ \displaystyle \sum_{k=1}^{l}\eta(1-\eta t)^{l-k} $ &
				$\eta\in(0,\kappa^2],\ \lambda=\frac{1}{\eta l}$ &
				any $\tau>0$ &
				$\bigl(1,\ \max\{1,(\tau/e)^{\tau}\}\bigr)$ \\
				
				\textbf{Spectral cutoff (SC)} &
				$ \displaystyle
				\begin{cases}
					1/t, & t\ge \lambda,\\
					0,   & \text{otherwise}
				\end{cases} $ &
				$\lambda>0$ &
				any $\tau>0$ &
				$(1,\,1)$ \\
				\hline
		\end{tabular}
	\end{threeparttable}
\end{table*}

Despite the theoretical appeal of spectral algorithms, their practical implementation can be computationally intensive, especially for large-scale datasets. The primary computational bottleneck arises from computing and storing the empirical covariance operator $C_{K,n}$, which involves handling an $n \times n$ kernel matrix (i.e., $n \times n$ matrix inverse in KRR). This results in significant time and space complexity (i.e., $O(n^3)$ and $O(n^2)$ for KRR), making it infeasible for large $n$. To alleviate this burden, various approximation techniques have been proposed to reduce the computational burden. One such technique is the use of random projections, which can effectively reduce the dimensionality of the data while preserving its essential structure \citep{vempala2005random}. The core idea of random projections is to restrict the original solution space to a lower-dimensional subspace using a projection operator $P$, satisfying $P=P^*=P^2$. Incorporating random projections into spectral algorithms leads to the following estimator:
\begin{align}\label{rp_spectral_estimator}
	f_{n,\lambda}^{rp}=Pg_{\lambda}(P C_{K,n}P)P S_{K,n}^*\mathbf{y}.
\end{align}
This approach can significantly reduce the computational complexity of spectral algorithms, making them more scalable and efficient for large datasets. In the next subsections, we will discuss some specific implementations of SARP, focusing on randomized sketches and Nystr\"{o}m sub-sampling.

\subsection{Randomized Sketches for Spectral Algorithms}\label{sec:SARP}

Randomized sketches involve a sketch matrix $\mathbf{G} \in \mathbb{R}^{m \times n}$ with $m \ll n$, and the range of the projection operator $P$ is defined as the closure of the span of the rows of $\mathbf{G}S_{K,n}$, i.e., $\text{Range}(P)=\overline{\text{span}\{S_{K,n}^*\mathbf{G}^T\}}$. The sketch matrix $G$ is drawn independently
of the training sample and is isotropically
normalized, i.e.,
$\mathbb E_G[G^\top G]=I_n$.
We assume that for any fixed nonzero
$\mathbf a\in\mathbb R^n$ and
$t\in(0,1)$, there exist constants
$C>0$ and $b\geq0$, independent of
$\mathbf a$, $m$, and $t$, such that
the following two-sided concentration
inequality holds:
\begin{equation}\label{sketch_property}
\change{\mathbb P_G\left(
\left|\|G\mathbf a\|_2^2-\|\mathbf a\|_2^2\right|
\geq t\|\mathbf a\|_2^2
\right)
\leq 2\exp\left(
-\frac{Cm t^2}{(\log n)^b}
\right),}
\end{equation}
Common choices for $\mathbf{G}$ include sub-Gaussian random sketches, randomized orthogonal system (ROS) sketches, and sub-sampling sketches \citep{yang2017randomized,lin2020convergences}.

\smallskip

\noindent \textbf{Sub-Gaussian random sketches.} A sketch matrix $\mathbf{G}\in\mathbb{R}^{m\times n}$ is called a normalized sub-Gaussian sketch if its entries $\{g_{ij}\}_{i,j=1}^{m,n}$ are independent, mean-zero random variables satisfying $\mathbb{E}[g_{ij}^{2}]=1/m$ and the sub-Gaussian condition $\mathbb{E}\exp(tg_{ij})\leq
\exp\left(c^{2}t^{2}/2m\right), t\in\mathbb{R},$ for some universal constant $c>0$.
The normalization ensures that
$\mathbb{E}[\mathbf{G}^{\top}\mathbf{G}]=I_n$.
Such sketches satisfy the two-sided concentration inequality \eqref{sketch_property} with $b=0$.
Typical examples include matrices with i.i.d. Gaussian entries distributed as $\mathcal{N}(0,1/m)$ or i.i.d. Rademacher (Bernoulli $\pm1$) entries scaled by $1/\sqrt{m}$.
Closely related constructions draw independent rows uniformly from a sphere of radius $\sqrt{n/m}$.
These rows are isotropic and sub-Gaussian and provide comparable sketching guarantees \citep{yang2017randomized}.

\smallskip

\noindent \textbf{Randomized orthogonal system (ROS) sketches.} A ROS sketch matrix $\mathbf{G} \in \mathbb{R}^{m \times n}$ is constructed as
$\mathbf{G}=\sqrt{n/m}\mathbf{RHD}$, where $\mathbf{D} \in \mathbb{R}^{n \times n}$ is a diagonal matrix with i.i.d. Rademacher entries, $\mathbf{H} \in \mathbb{R}^{n \times n}$ is an orthonormal matrix (e.g., Hadamard or Fourier matrix), and $\mathbf{R} \in \mathbb{R}^{m \times n}$ is a random sampling matrix that selects $m$ rows uniformly at random from the $n \times n$ identity matrix. ROS sketches are computationally efficient due to the fast multiplication algorithms available for orthonormal matrices like Hadamard and Fourier transforms. Typically, a matrix-vector multiplication with a ROS sketch can be performed in $O(n \log m)$ time, which is significantly faster than the $O(mn)$ time required for dense random matrices \citep{ailon2009fast}.

\smallskip

\noindent \textbf{Sub-sampling sketches.} A sub-sampling sketch matrix $\mathbf{G} \in \mathbb{R}^{m \times n}$ is obtained by selecting $m$ rows uniformly at random without replacement from the $n \times n$ identity matrix. This means that each row of $\mathbf{G}$ contains exactly one entry equal to $\sqrt{n/m}$, and all other entries are zero. Unlike normalized sub-Gaussian and ROS
sketches, coordinate subsampling does
not generally satisfy \eqref{sketch_property} for arbitrary vectors. Its projection error is therefore analyzed separately using Nystr\"{o}m
subsampling bounds.

\subsection{Nystr\"{o}m Sub-sampling for Spectral Algorithms}\label{sec:Nystron}

\textbf{Plain Nystr\"{o}m.} Plain Nystr\"{o}m sub-sampling can be viewed as a sub-sampling sketch using matrix $\mathbf{G}$, where the range of the projection operator $P$ is defined as the closure of the span of a subset of the training data points. Specifically, we randomly and uniformly select a subset of $m$ data points $\{\bfx_{i_j}\}_{j=1}^m$ from the training dataset $\{\bfx_i\}_{i=1}^n$ without replacement, and define the range of $P$ as $\text{Range}(P)=\overline{\text{span}\{K_{\bfx_{i_j}}\}_{j=1}^m}$. 

\smallskip

\noindent \textbf{Approximate leverage score (ALS) Nystr\"{o}m.} 
\citet{alaoui2015fast} proposed the leverage score sampling for Nystr\"{o}m sub-sampling, which corresponds to a non-uniform sub-sampling sketch. The leverage scores quantify the importance of each data point in the dataset:
\[ (l_i(\lambda))_{i=1}^n,\quad l_i(\lambda)=\bigl(\mathbf{K}(\mathbf{K}+\lambda I)^{-1}\bigr)_{ii}, \quad i=1,2,\ldots,n.\]
where $\mathbf{K}=S_{K,n}S_{K,n}^*$ denotes the Gram matrix. 
In practice, exact leverage scores are often replaced by efficiently computable approximations $\{\hat{l}_i(\lambda)\}_{i=1}^n$ \citep{musco2017recursive,calandriello2017distributed,rudi2018fast,chen2021fast} that satisfy
\begin{align*}
\hat{l}_i(\lambda)/L \leq l_i(\lambda) \leq L\hat{l}_i(\lambda) \quad \hbox{for some $L \geq 1$.}
\end{align*}
The ALS Nystr\"{o}m method selects $m$ data points $\{\widetilde{\bfx}_{i_j}\}_{j=1}^m$ from the training dataset $\{\bfx_i\}_{i=1}^n$ according to the probability distribution $\{q_i\}_{i=1}^n$, where
$q_i=\hat{l}_i(\lambda)/\sum_{j=1}^n \hat{l}_j(\lambda)$. The range of the associated projection operator $P$ is then defined by $\text{Range}(P)=\overline{\text{span}\{K_{\widetilde{\bfx}_{i_j}}\}_{j=1}^m}$.

To measure the computational complexity of the Nystr\"{o}m method under different sampling strategies, \citet{alaoui2015fast} introduced two empirical effective dimensions:
\begin{align*}
  d_{\infty}(\lambda)=n\|\operatorname{diag}(\mathbf{K}(\mathbf{K}+ \lambda I)^{-1})\|_{\infty},  \quad
  d_{\text{eff}}(\lambda)=\operatorname{Tr}\bigl(\mathbf{K}(\mathbf{K}+ \lambda I)^{-1}\bigr)=\sum_{i=1}^nl_i(\lambda).
\end{align*}
For any $\lambda>0$, it follows directly that $d_{\text{eff}}(\lambda)  \leq d_{\infty}(\lambda) \leq \kappa^2/\lambda$. \citet{bach2013sharp} and \citet{alaoui2015fast} further showed that the minimum number of Nystr\"{o}m samples required to achieve accurate approximation scales proportionally to $d_{\infty}(\lambda)$ for plain Nystr\"{o}m, and to $d_{\text{eff}}(\lambda)$ 
when approximate leverage score Nystr\"{o}m is employed.

\section{Main Results}\label{sec:main}

In this section, we first establish a minimax lower bound for the learning rate in Sobolev norms under general source conditions, which offers the fundamental learning limits of SARP. Then we show that the upper bound for the learning rate of SARP can match the lower bound under mild conditions. The analysis relies on the following definitions and assumptions. 

\begin{assumption}[\textbf{Noise condition}]\label{ass1}
There exist constants $\sigma\geq0$ and $M\geq0$ such that for all $p\geq 2$, $\mathbb{E}(|\varepsilon|^{p}| \bfx)\leq \frac{1}{2}p!\sigma^{2}M^{p-2}.$ 
\end{assumption}

This assumption is the classical Bernstein condition imposed on the noise, which is widely used in kernel learning theory \citep{caponnetto2007optimal,rastogi2017optimal}. It is satisfied, for example, by bounded, Gaussian, or sub-Gaussian noise. In contrast, the closely related work \citep{lin2020convergences} assumed that the regression function $f_{\rho}$ is uniformly bounded. Such a requirement can be overly restrictive, and a uniform bound on $f_\rho$ is very hard to verify from finite data in practice. Assumption~\ref{ass1} instead places the Bernstein condition solely on the noise $\varepsilon$ and does not require $f_{\rho}$ to be uniformly bounded. In our synthetic experiment in Section \ref{app:numerical}, the regression function diverges when $s=0.5$; hence $f_\rho$ is not uniformly bounded, yet SARP still achieves the optimal Sobolev learning rates. This confirms that a uniform boundedness assumption is unnecessary for obtaining optimal learning rates for SARP. \change{We acknowledge Assumption \ref{ass1} still imposes a relatively strong light-tailed requirement on the noise, which limits the applicability of our results to heavy-tailed settings. We also remark that recent work has established KRR theory under heavy-tailed noise with finitely many bounded moments \citep{mollenhauer2025regularized}. Adapting our framework to this setting requires a separate and careful analysis, which we leave as promising future work. Further discussion on this extension is given in Section IV-A of the Supplementary Material.}

\begin{definition}[\textbf{Effective dimension}]\label{def_effective_dimension}
    For $\lambda >0$, define the random variable ${\cal N}_{\bfx}(\lambda)=\langle K_{\bfx}, (C_K+\lambda I)^{-1}K_{\bfx}\rangle_{\cal H}$ with $\bfx \in {\cal X}$ drawn from $\rho_{\cal X}$. Then the average effective dimension and maximal effective dimension of RKHS ${\cal H}_K$ are defined as ${\cal N}(\lambda)=\mathbb{E}{\cal N}_{\bfx}(\lambda)$ and ${\cal N}_{\infty}(\lambda)=\operatorname*{ess} \sup_{\bfx\in X}{\cal N}_{\bfx}(\lambda)$, respectively.
\end{definition}

The effective dimension characterizes the complexity of the RKHS ${\cal H}_K$ and plays a crucial role in learning theory \citep{caponnetto2007optimal,steinwart2009optimal,guo2017learning}. It is easy to see that ${\cal N}(\lambda)=\operatorname{Tr}((C_K+\lambda I)^{-1}C_K)=\sum_{i=1}^{\infty}\frac{\mu_i}{\mu_i+\lambda}$ and ${\cal N}(\lambda)\leq {\cal N}_{\infty}(\lambda)\leq \kappa^2\lambda^{-1}$, where $\operatorname{Tr}(\cdot)$ denotes the trace of the operator. 

\begin{assumption}[\textbf{Capacity condition}]\label{ass2}
    For $\lambda>0$, there exist constants $Q>0$ and $\beta\in (1,\infty)$ such that ${\cal N}({\lambda}) \leq Q^2\lambda^{-\frac{1}{\beta}}.$ 
\end{assumption}

This assumption is known as the capacity condition. It can also be verified by the decay rate of the eigenvalues $\{\mu_i\}_{i\geq 1}$. For example, if $\mu_i \leq c i^{-\beta}$ for some $c>0$ and $\beta > 1$, then ${\cal N}(\lambda) \leq c_{\beta}\lambda^{-\frac{1}{\beta}}$ for some constant $c_{\beta}>0$.

\begin{assumption}[\textbf{Embedding condition}]\label{ass3}
Assume that there exists an order $\alpha \in [1/\beta,1]$ and a constant $A>0$ such that the canonical inclusion map $i^\alpha : [{\cal H}_K]_{\rho}^{\alpha} \hookrightarrow L_\rho^\infty({\cal X}), i^\alpha(f)=f$,
is bounded, i.e. $
\|f\|_{\infty} 
  \le A \,\|f\|_{[\cal H]_\rho^\alpha},
   \forall f \in [{\cal H}_K]_{\rho}^{\alpha}.$
Equivalently, $[{\cal H}_K]_\rho^\alpha$ is continuously embedded into $L_\rho^\infty(\cal X)$ and the operator norm of the embedding satisfies $\|i^\alpha\|\le A$. We also define the embedding index as $\alpha_{0}=\inf \left\{\alpha \in\left[1/\beta, 1\right]:\left\|\mathcal{H}|^{\alpha} \hookrightarrow L^{\infty}(X)\right\|<\infty\right\}$ and let $\alpha_0 \in [1/\beta,1)$.
\end{assumption}

\change{The embedding property indicates that the interpolation space $[{\cal H}]_{\rho}^{\alpha}$ can be continuously embedded into $L^{\infty}_{\rho}({\cal X})$, with operator norm bounded by $A$. By the inclusion relationship of interpolation spaces, we have ${\cal H}_K$ has the embedding property of order $\alpha^{\prime}$ whenever it has the embedding property of order $\alpha$ for any $\alpha^{\prime} \geq \alpha$. Typically, a larger $\alpha$ reflects a weaker embedding. The condition holds trivially when $\alpha=1$ since the kernel $K$ is bounded. The case $\alpha_0=1/\beta$ corresponds to the benign RKHSs, in which the feature space is nearly homogeneous and contains no extreme leverage points or spatial spikes. Such benign cases can be referred to in Section III of the Supplementary Material. However, the embedding condition of order $\alpha_0$ itself may not hold.}

 \begin{remark}
\change{Assumption \ref{ass3} yields the maximal capacity condition, i.e., ${\cal N}_{\infty}(\lambda) \leq A^2\lambda^{-\alpha}$, which is important to our analysis in the mis-specified setting. We provide the detailed proof in Lemma 11 of the Supplementary Material. However, the converse fails in general, and the embedding assumption is strictly stronger. In our theoretical framework, deriving a precise bound for the $L^\infty$-norm is a critical step. The maximal capacity condition alone is insufficient as it cannot decouple the spatial supremum from the eigenvalue decay. The embedding assumption resolves this issue directly by establishing the required inequality, i.e., $\|f\|_{\infty}\leq A\|L_{K}^{-\alpha/2}S_Kf\|_{\rho}$ for any $f \in {\cal H}_K$.
Therefore, the embedding condition is mathematically required for obtaining the required tight bound. Additional discussion and counterexamples appear in Section V-D of the Supplementary Material.} 
\end{remark}

\begin{assumption}[\textbf{General source condition}]\label{ass4}
       For a fixed source index function
$\phi\in\mathcal C_s$ and $R>0$, where $C_s$ is the index function class defined in \eqref{cs_class}, assume that
        \begin{align}\label{ass4_e1}
\hspace{-10mm}         f_{\rho}\in \Omega_{\phi, s, R}:=\left\{f \in L^2_{\rho_{\cal X}}: f=\phi(L_K)g, \|g\|_{\rho}\leq R \right\}.
        \end{align}
 For $\lambda \in (0, \kappa^2]$, we further assume that there exists $c>0$ such that $c\lambda^{\tau}/\phi(\lambda)\leq \inf_{t \in [\lambda, \kappa^2]}t^{\tau}/\phi(t)$.
\end{assumption}

A typical example is the H\"{o}lder source condition, i.e., $\phi(t)=t^{s/2}$ for some $s>0$. In this case, $f_{\rho} \in \Omega_{\phi,s,R}$ implies that $f_{\rho} \in [{\cal H}_K]_{\rho}^{s}$ and $\|f_{\rho}\|_{[{\cal H}]_{\rho}^s}\leq R$, as also considered in \citet{fischer2020sobolev,zhang23mis,zhang2024optimality}. The parameter $s$ quantifies the smoothness of $f_{\rho}$ relative to the RKHS ${\cal H}_K$. Specifically, if $s>1$, then $f_{\rho}$ is smoother than the functions in ${\cal H}_K$ ($f_{\rho} \in {\cal H}_K$), whereas $0<s<1$ means $f_{\rho}$ is less smooth than the functions in ${\cal H}_K$ ($f_{\rho} \in [{\cal H}_K]_{\rho}^{s}$). Practically, one often employs a very smooth kernel (Gaussian or high-smoothness Matérn kernel) while the true regression function possesses only finite Sobolev regularity. Such "kernel too smooth, truth less smooth" situations are common, for instance, a Gaussian kernel is used as a default and still yields good empirical performance. Our theory will show that this is reasonable since SARP can still achieve the optimal rates in this mis-specified setting. In \change{ Corollary \ref{cor1}}, we further present the Hölder-source rates in three smoothness regimes.

\subsection{Minimax Lower Bound for Sobolev Norm Learning}\label{sec:minimax_lower_bound}

In order to establish the lower bound, we introduce the following additional assumption, which can be viewed as an enhanced version of Assumption \ref{ass2}. Analogous assumptions appear in \citet{fischer2020sobolev,zhang2024optimality}.

\begin{assumption}[\textbf{Eigenvalue decay condition}]\label{ass2_}
Assume that there exist some positive constants $0<q_1\leq q_2$ and $\beta\in (1,\infty)$ such that
$q_1 k^{-\beta} \leq \mu_k \leq q_2 k^{-\beta},   \forall k\geq 1$, where $\mu_k$ is the $k$-th eigenvalue of the integral operator $L_K$. Thus, it also holds that ${\cal N}(\lambda)\asymp \lambda^{- {1}/{\beta}}$.
\end{assumption}

\begin{theorem}\label{low_thm}
Let the constants $\sigma, M, R, Q$ be fixed. Given $s>0$ and $\beta \in (1,\infty]$, let ${\mathcal P}={\mathcal P}(\phi,s,\beta)$ be the set of all probability distributions $\rho$ such that Assumptions \ref{ass1}, \ref{ass2_} and \eqref{ass4_e1} hold. Then for any $\gamma \in [0, s\wedge 1]$, $\delta \in (0,3/8)$, and all possible estimators $\widehat{f}$ based on the $n$ i.i.d. samples, there exists a probability measure $\rho^* \in \mathcal{P}$ such that  with probability at least $1-\delta$, 
\begin{align}\label{low_thm_eq1}
\|\widehat{f}-f_{\rho^*}\|_{[{\cal H}]_{\rho^*}^{\gamma} }^2 \geq C\delta\phi^2(\Phi^{-1}(n^{-1}))(\Phi^{-1}(n^{-1}))^{-\gamma},
\end{align}
where $\Phi(t)=(\phi(t)/\phi(1))^2t^{1/\beta}$, $\Phi^{-1}(\cdot)$ denotes the functional inverse, and $C$ is a positive constant independent of $\lambda$, $n$, $\delta$, whose explicit expression is given in the proof. 
\end{theorem}

Theorem \ref{low_thm} establishes a minimax lower bound for the learning rate in Sobolev norms under general source conditions.  In the existing literature, minimax lower bounds for the $L^2_{\rho}$ norm or $\cal H$ norm are typically derived for the well-specified case where $f_{\rho} \in {\cal H}_K$ \citep{caponnetto2007optimal,rastogi2017optimal}, which corresponds to the source condition with $s \geq 1$. For the mis-specified case where $f_{\rho} \notin {\cal H}_K$, \citet{fischer2020sobolev,zhang23mis,zhang2024optimality} provide the minimax lower bounds with $s>0$. However, these studies all assume the Hölder source condition, which is a special case of the more general source condition considered in this paper. To the best of our knowledge, this is the first work to establish minimax lower bounds under general source conditions for both well-specified and mis-specified cases. Furthermore, our analysis does not rely on the uniform boundedness assumption on $f_{\rho}$ and accounts for multiple Sobolev norms. When $\phi(t)=t^{s/2}$ for some $s>0$, the bound in Theorem~\ref{low_thm} recovers the existing lower bounds of \citet{fischer2020sobolev,zhang23mis,zhang2024optimality}, as stated in the following corollary. \change{We also want to emphasize that the minimax lower bound in Theorem \ref{low_thm} does not rely on the qualification parameter $\tau$ and therefore does not require any restriction on $s$. So the lower bound is valid for any $s>0$.}

\begin{corollary}\label{low_cor}
Under the same conditions of Theorem \ref{low_thm}, if $\phi(t)=t^{s/2}$ for some $s>0$, then with probability at least $1-\delta$, there holds that
\begin{align}\label{low_cor_eq1}
\|\widehat{f}-f_{\rho^*}\|_{[{\cal H}]_{\rho^*}^{\gamma} }^2 \geq C\delta n^{-\frac{(s-\gamma)\beta}{s\beta+1}}.
\end{align}
\end{corollary}

\begin{remark}
\change{The lower bound in Theorem \ref{low_thm} applies to standard spectral algorithms without random projections. Deriving a corresponding bound that explicitly incorporates the projection error is technically challenging and such lower bounds remain largely open. Nevertheless, Theorem \ref{low_thm} serves as a benchmark for the optimal rate without computational constraints. Our upper bounds explicitly quantify the projection effect through the projection dimension and the approximation error $\mathcal A_{K,P}$, showing when the projected estimators match the minimax rate.  Finally, since $f_{n,\lambda}^{rp}\in\operatorname{Ran}(P)$, we have
$$
\|f_{n,\lambda}^{rp}-f_\rho\|_{[\mathcal H]_\rho^\gamma}^2
\geq
\inf_{f\in\operatorname{Ran}(P)}
\|f-f_\rho\|_{[\mathcal H]_\rho^\gamma}^2.
$$
This inequality provides a projection-dependent lower bound that measures how well $f_\rho$ can be approximated within the projected subspace. In particular, the approximation error can be zero when $f_\rho$ belongs to $\operatorname{Ran}(P)$, even if $P\neq I$. Therefore, deriving a sharp minimax lower bound that explicitly accounts for random projections would require additional assumptions on the relationship between $f_\rho$ and the projected subspace. We leave this question for future investigation.}
\end{remark}

\subsection{Learning Rates for General Random Projections}

In this section, we present our main theoretical results on the learning rates of SARP. For clarity, we introduce a projection operator term ${\cal A}_{K,P}= \|C_K^{1/2}(I-P)\|$ and the projection error $\Delta$, which will be specified in the theorem below. Throughout the analysis, we assume $s\in (0,2\tau]$. \change{This range follows from the qualification of the spectral filter and is needed for the upper-bound analysis.} Recall that $f_{n,\lambda}^{rp}$ is given in \eqref{rp_spectral_estimator}. For ease of reference, we summarize the technical parameters used in the main results in Table \ref{tab:key_concepts}.

\begin{table*}[t]

    \centering

    \caption{\change{Summary of technical parameters used in the main results.}}

    \label{tab:key_concepts}

    \begin{tabular}{@{}c l p{10cm}@{}}

        \toprule

        \textbf{Parameter} & \textbf{Concept} & \textbf{Intuitive Meaning} \\

        \midrule

        $\beta$ & Capacity condition & Measures the complexity of the RKHS (a larger $\beta$ indicates a smaller space). \\

        $\alpha$ & Embedding condition & Controls the continuous embedding of the RKHS interpolation space into $L^\infty$ (a larger $\alpha$ reflects a weaker embedding).\\

        $\alpha_0$ & Embedding index & The smallest embedding exponent in the limiting sense.\\

        $s$ & Source condition & Characterizes the smoothness of target function (a larger $s$ implies a smoother function). \\

        $\gamma$ & Norm index & Determines the error evaluation norm ($\gamma=0$ for standard $L^2$ error, $\gamma=1$ for RKHS error). \\

        \bottomrule

    \end{tabular}

\end{table*}

\begin{theorem}\label{thm1}
Under Assumptions \ref{ass1}-\ref{ass4}, for any $\gamma \in [0,1 \wedge s]$, $\lambda\in (0,1]$ and $\delta \in (0,1)$, if $n\gtrsim \lambda^{-\alpha}\log(32\kappa^2/\lambda\delta)\vee D_{q,\delta}^{\frac{2\alpha}{s+\alpha}}
\left(\lambda^\alpha\phi^2(\lambda)\right)^{-\frac{\alpha}{s+\alpha}}$ where $D_{q,\delta}$ is defined in the proof, then with probability at least $1-\delta$, the following holds:

\noindent (1) If $\phi: [0,\kappa^2]\rightarrow \mathbb{R}^+$ is operator monotone with $\phi(0)=0$ and $\phi(\kappa^2)< \infty$, or $\phi:[0,b]\to\mathbb R$ is real analytic on an open neighborhood of $[0,b]$ and
satisfies $\phi(0)=0$, then
\begin{align*}
\|f_{n,\lambda}^{rp}-f_{\rho}\|_{[{\cal H}]_{\rho}^{\gamma}}  \lesssim   \lambda^{-\frac{\gamma}{2}}\bigg[\phi(\lambda)+\frac{1}{n\lambda^{\max\{\frac{\alpha}{2},\alpha-\frac{s}{2}\}}} +\frac{1}{\sqrt{n}\lambda^{\frac{1}{2\beta}}}+\frac{\phi(\lambda)}{\sqrt{n}\lambda^{\frac{\alpha}{2}}}+\Delta\bigg]\log\frac{8}{\delta}.
\end{align*}

\noindent (2) If $\phi=\psi\vartheta$, where $\psi:[0,b]\to\mathbb R_+$ is operator monotone, where $b>2\kappa^2$,
$\psi(0)=0$, and $\psi(b)<\infty$, and $\vartheta: [0,\kappa^2]\rightarrow \mathbb{R}^+$ is non-decreasing, real analytic on an open neighborhood of $[0,b]$ and
satisfies $\vartheta(0)=0$, and for $\forall \lambda \in (0, \kappa^2]$, there exists $c^{\prime}>0$ such that $ c^{\prime}\frac{\lambda^{\tau}}{\vartheta(\lambda)\lambda^{1/2}}\leq \inf_{t \in [\lambda, \kappa^2]}\frac{t^{\tau}}{\vartheta(t)t^{1/2}}$, then
\begin{small}
\begin{align*}
\|f_{n,\lambda}^{rp}-f_{\rho}\|_{[{\cal H}]_{\rho}^{\gamma}} 
\lesssim   \lambda^{-\frac{\gamma}{2}}\bigg[\phi(\lambda)+\frac{1}{n\lambda^{\max\{\frac{\alpha}{2},\alpha-\frac{s}{2}\}}}+\frac{1}{\sqrt{n}\lambda^{\frac{1}{2\beta}}}  +\frac{\phi(\lambda)}{\sqrt{n}\lambda^{\frac{\alpha}{2}}}+\vartheta(\lambda)\psi\bigg(\frac{1}{\sqrt{n}}\bigg)+\Delta\bigg]\log\frac{8}{\delta}.
\end{align*}
\end{small}
Here, the projection error 
$\Delta = \begin{cases}  \lambda^{\frac{s}{2}-1}({\cal A}_{K,P}^2+\lambda^{\frac{1}{2}}{\cal A}_{K,P}+\lambda^{\frac{\gamma}{2}}{\cal A}_{K,P}^{2-\gamma})+\lambda+{\cal A}_{K,P}^2, &\hspace{-3mm}  \text { if } s < 2; \\ {\cal A}_{K,P}^2+n^{-\frac{1}{4}}{\cal A}_{K,P}+\lambda^{\frac{\gamma}{2}}{\cal A}_{K,P}^{2-\gamma}+n^{-\frac{1}{2}}, & \hspace{-3mm} \text { if } s \geq 2. \end{cases}$
\end{theorem}

Theorem \ref{thm1} provides high-probability Sobolev norm error bounds for the SARP under general source conditions. Here we consider three types of source conditions: (1) a non-decreasing index function $\phi$; (i.e., the power index function or the Hölder source condition, $\phi(t)=t^{s/2}$) (2) a real-analytic index function $\phi$; (3) a composite index function $\phi=\psi\vartheta$. Compared to the general source conditions proposed in \citet{bauer2007regularization,lin2020optimal}, our main contribution lies in the refined condition imposed on the index function, which is carefully exploited to achieve tight error bounds. Specifically, we propose a new lemma providing a sharp bound for $\|P\phi(C_{K,n})P-\phi(PC_{K,n}P)\|$ using ${\cal A}_{K,P}= \|C_K^{1/2}(I-P)\|$. To the best of our knowledge, this lemma and its proof have not appeared before. Our proof is based on a delicate block-operator decomposition of $C_{K,n}$ together with the resolvent identity and Cauchy's integral formula of $\phi$, rather than standard spectral perturbation or polynomial-truncation arguments, and thus appears to be genuinely new.

The derived learning rates can be decomposed into several terms, each reflecting different aspects of the learning process. The term $\lambda^{-\frac{\gamma}{2}}\phi(\lambda)$ represents the approximation error, which depends on the regularization parameter $\lambda$ and the smoothness of the regression function $f_{\rho}$ characterized by the index function $\phi$. The intermediate terms involving $n$, $\lambda$, $\beta$, and $\alpha$ collectively describe the empirical error, which captures the effect of finite-sample variance. The last term $\Delta$ accounts for the impact of random projections on learning performance. When $P=I$, our result recovers the existing bounds in \citet{lin2020optimal} for spectral algorithms without RP. To achieve the optimal trade-off between these error terms,  the regularization parameter $\lambda$ must be chosen appropriately, which leads to the following theorem.

\begin{theorem}\label{thm2}
Under the same assumptions and conditions of Theorem \ref{thm1}, let $s>\alpha_0-1/\beta$, and $\lambda=\Phi^{-1}(n^{-1})$ with $\Phi(t)=(\phi(t)/\phi(1))^2t^{1/\beta}$ and $\Phi^{-1}(\cdot)$ denoting the functional inverse. For case (2), we additionally let $\lambda\geq n^{-1/2}$, then with probability at least $1-\delta$, there holds that
\begin{align*}
\|f_{n,\lambda}^{rp}-f_{\rho}\|^2_{[{\cal H}]_{\rho}^{\gamma}} \lesssim (\phi(\Phi^{-1}(n^{-1}))+\Delta)^2(\Phi^{-1}(n^{-1}))^{-\gamma}\log^2\frac{8}{\delta},
\end{align*}
where the projection error $\Delta$ is the same as that in Theorem \ref{thm1}.
\end{theorem}

Compared to the lower bound in Theorem \ref{low_thm}, Theorem \ref{thm2} shows that SARP attains the minimax-optimal learning rates up to a logarithmic factor whenever the projection error $\Delta$ is of a smaller order than $\phi(\Phi^{-1}(n^{-1}))$. This requirement can be satisfied by choosing an appropriate projection dimension $m$ for each type of random projection, as discussed in the next section. When $P=I$, the resulting learning rates align with those in \citet{bauer2007regularization} and \citet{lin2020optimal} for spectral algorithms under general source conditions. However, our results are more general as they encompass a broader range of source index $s>\alpha_0-1/\beta$ with $\alpha_0 \in [1/\beta,1)$ and Sobolev norms $\gamma \in [0,1 \wedge s]$, while the previous works primarily focus on the well-specified case with $s\geq 1$ \citep{bauer2007regularization} or part of the mis-specified case $s\geq 1-1/\beta$ \citep{lin2020optimal}. 

\begin{remark}
\change{The Sobolev norm provides a more general measure of
learning performance than the traditional $L^2_{\rho_X}$
norm. For the squared loss, we have
\[
\mathcal E(f)-\mathcal E(f_\rho)
=\|f-f_\rho\|_{\rho}^2
=\|f-f_\rho\|_{[\mathcal H]_\rho^0}^2.
\]
Thus, setting $\gamma=0$ in Theorems \ref{thm1} and \ref{thm2} directly yields the corresponding population excess risk bounds. In particular, when $s>\alpha_0-1/\beta$, SARP achieves the minimax-optimal excess risk rate
up to logarithmic factors, provided that the projection error is sufficiently small. For $\gamma>0$, the Sobolev norm provides stronger information by taking into account the smoothness of the estimated function. In particular, for Sobolev-type RKHSs, these norms can measure errors in fractional-order smoothness. For example, when
$K(x,x')=\min\{x,x'\}$ and $\gamma=1$, we have
\[
\|f-f_\rho\|_{[\mathcal H]_\rho^1}^2
=\int_0^1 |f'(x)-f_\rho'(x)|^2\,dx,
\]
provided that $f,f_\rho\in\mathcal H_K$.
This allows us to evaluate the recovery of derivatives,
which cannot be directly measured by the population
excess risk.}
\end{remark}

\subsection{Learning Rates for Specific Random Projections}\label{sec:applications}
In this section, we apply Theorem \ref{thm2} to three specific types of random projections introduced earlier: randomized sketches, plain, and ALS Nystr\"{o}m sub-sampling. We show how to choose the projection dimension $m$ to effectively control the projection error $\Delta$, thereby achieving optimality. For simplicity, we focus on the Hölder source condition with $\phi(t)=t^{s/2}$ in the following corollary. 

\begin{corollary}\label{cor1}
Under the same assumptions and conditions of Theorem \ref{thm1}, let $\phi(t)=\kappa^{-(s-2)_{+}}t^{{s}/{2}}$ for some $s>0$, and $n$ is sufficiently large. We consider the following three cases of $s$:

\noindent (1) When $s\leq \alpha_0- {1}/{\beta}$, for any $\alpha>\alpha_0$, let $\lambda \asymp \left(\frac{C\log n}{n}\right)^{\frac{1}{\alpha}}$ for sufficiently large $C$,  and the projection dimension $m$ satisfies
\begin{align*}
  m \gtrsim \begin{cases} n^{\frac{1}{\alpha\beta}}, & \text {for Randomized Sketches and ALS Nystr\"{o}m}; \\ n, & \text { for Plain Nystr\"{o}m}. \end{cases}
\end{align*}
Then with probability at least $1-\delta$, there holds that
\begin{align*}
\|f_{n,\lambda}^{rp}-f_{\rho}\|^2_{[{\cal H}]_{\rho}^{\gamma}} \lesssim n^{-\frac{s-\gamma}{\alpha}}(\log n)^{\frac{s-\gamma}{\alpha}}\log^2\frac{10}{\delta}.
\end{align*}

\noindent (2) When $s> \alpha_0- {1}/{\beta}$ and $s<2$,  let $\lambda\asymp n^{-\frac{\beta}{s\beta+1}}$, and the projection dimension $m$ satisfies 
\begin{align*}
  m \gtrsim \begin{cases} n^{\frac{1}{s\beta+1}}, & \text {for Randomized Sketches and ALS Nystr\"{o}m}; \\ n^{\frac{\alpha\beta}{s\beta+1}}, & \text { for Plain Nystr\"{o}m}. \end{cases}
\end{align*}
Then with probability at least $1-\delta$, there holds that
\begin{align*}
\|f_{n,\lambda}^{rp}-f_{\rho}\|^2_{[{\cal H}]_{\rho}^{\gamma}} \lesssim n^{-\frac{(s-\gamma)\beta}{s\beta+1}}\log^2\frac{10}{\delta}.
\end{align*}

\noindent (3) When $s\geq 2$, let $\lambda \asymp n^{-\frac{\beta}{s\beta+1}}$, and the projection dimension $m$ satisfies 
\begin{align*}
  m \gtrsim \begin{cases} n^{\frac{s-\gamma}{(2-\gamma)(s\beta+1)}}, & \text {for Randomized Sketches and ALS Nystr\"{o}m}; \\ n^{\frac{\alpha\beta(s-\gamma)}{(2-\gamma)(s\beta+1)}}, & \text { for Plain Nystr\"{o}m}. \end{cases}
\end{align*}
Then with probability at least $1-\delta$, there holds that
\begin{align*}
\|f_{n,\lambda}^{rp}-f_{\rho}\|^2_{[{\cal H}]_{\rho}^{\gamma}} \lesssim n^{-\frac{(s-\gamma)\beta}{s\beta+1}}\log^2\frac{10}{\delta}.
\end{align*}

\end{corollary}

Corollary \ref{cor1} provides capacity-dependent learning rates for spectral algorithms with different types of random projections under the Hölder source condition. The results demonstrate that by selecting an appropriate projection dimension $m$ proportional to the average or maximal effective dimension, the projection error $\Delta$ can be controlled to be of a smaller order than the approximation and empirical error, thereby achieving the minimax optimal learning rates $n^{-(s-\gamma)\beta/(s\beta+1)}$ up to a logarithmic factor when $s> \alpha_0-1/\beta$. The learning rates degrade to $n^{-(s-\gamma)/\alpha}$ when $s\leq \alpha_0-1/\beta$ and $\alpha>\alpha_0$. In the most benign case when $\alpha_0=1/\beta$, the optimal learning rates are achieved for all $s\in (0,2\tau]$. For ease of notation, we show
only the polynomial orders of the
projection dimensions in Corollary \ref{cor1}. The full sufficient conditions including all logarithmic factors are stated in Lemmas 21-23 of the Supplementary Material.

Compared with the existing results for spectral algorithms (or KRR) with randomized  projection under the Hölder source condition, our guarantees extend to the mis-specified case with $s\leq \alpha_0-1/\beta$ and Sobolev norms $\gamma \in [0,1 \wedge s]$. For example, \citet{yang2017randomized} only considered the worst case with $s=1$ and $\gamma=0$ under fixed design; \citet{rudi2015less} established the optimal learning rates solely in the well-specified case with $s\geq 1$ and $\gamma=0$; the state-of-the-art results \citep{lin2020convergences} established learning rates in various norms but only for part of the mis-specified case, requiring $s\geq 1-1/\beta$. Our extension comes from tighter estimates of (i) the similarity between the population covariance operator $C_K$ and its empirical counterpart $C_{K,n}$, which relaxes the sample‑size requirement from $n \gtrsim \lambda^{-1}$ to $n \gtrsim \lambda^{-\alpha}$, and (ii) the sample-variance term, reducing its order from ${\cal O}(n^{-1}\lambda^{\min\{-1/2,\, s/2-1\}}+n^{-1/2}\lambda^{(s-1)/2})$ to ${\cal O}(n^{-1}\lambda^{\min\{-\alpha/2,\, s/2-\alpha\}}+n^{-1/2}\lambda^{(s-\alpha)/2})$. 
Furthermore, different from the analysis in \citet{lin2020convergences}, we do not impose the uniform boundedness of $f_{\rho}$, which has been discussed in Assumption \ref{ass1}.

\begin{remark}
\change{The embedding condition yields the maximal capacity condition, i.e., ${\cal N}_{\infty}(\lambda) \lesssim \lambda^{-\alpha}$ for any $\alpha>\alpha_0$, which determines the sample-size requirement $n\gtrsim\lambda^{-\alpha}$ up to logarithmic factors. In Theorem \ref{thm1}, the upper bounds explicitly rely on the embedding index $\alpha$ through the maximal capacity condition. For the Hölder source condition in Corollary \ref{cor1}, balancing the approximation and stochastic errors yields $\lambda \asymp n^{-\beta/(s\beta+1)}$ with $s>\alpha_0-1/\beta$, which in turn gives the final learning rate $n^{-s\beta/(s\beta+1)}$ up to logarithmic factors. The choice of $\lambda$ also satisfies the embedding-dependent sample-size requirement $n\gtrsim \lambda^{-\alpha}$ when $s>\alpha_0-1/\beta$ and $n$ is sufficiently large. Consequently, although the final optimal learning rate appears independent of the embedding index $\alpha$, the sample-size requirement implicitly depends on it, reflecting the influence of the embedding condition on the overall analysis. However, when $s\leq \alpha_0-1/\beta$, the embedding condition becomes the limiting factor, leading to the slower rate $(\log n/n)^{(s-\gamma)/\alpha}$ with $\lambda\asymp (\log n/n)^{1/\alpha}$. Thus, a smaller embedding $\alpha$ enlarges the range in which minimax-optimal rates can be achieved. Moreover, $\alpha$ influences the projection dimension required for the learning rates, as shown in Corollary \ref{cor1}.}    
\end{remark}

\begin{remark}\label{rem1}
Ignoring the logarithmic terms, for randomized sketches and ALS Nyström methods, we can establish a sharp bound on ${\cal A}_{K,P}$ provided $m \gtrsim \lambda^{-{1}/{\beta}}$. For the plain Nyström method, it suffices that $m \gtrsim \lambda^{-{\alpha}}$. These requirements follow directly from Assumption \ref{ass2} that the average effective dimension satisfies ${\cal N}(\lambda) \lesssim \lambda^{-1/\beta}$, and by Assumption \ref{ass3}, the maximal effective dimension satisfies ${\cal N}_{\infty}(\lambda) \lesssim \lambda^{-\alpha}$. The conditions above can be equivalently expressed as $m \gtrsim {\cal N}(\lambda)$ for randomized sketches and ALS Nyström; $m \gtrsim {\cal N}_{\infty}(\lambda)$ for plain Nyström. In the proof of Corollary \ref{cor1}, we explicitly obtain the optimal rates by choosing $m$ so that the projection term does not exceed the bias and variance terms under these conditions. Consequently, the sufficient projection dimension is proportional to the (average or maximal) effective dimension to guarantee the optimal rates.
\end{remark}

\begin{remark}\label{rem3}
Computationally,  randomized sketches and ALS Nystr\"{o}m require a projection dimension $m$ that is significantly smaller than the sample size $n$, yielding substantial computational savings. In contrast, plain Nystr\"{o}m demands a larger projection dimension, which may limit its computational efficiency in certain scenarios ($s\leq \alpha_0-1/\beta$). Compared to the optimal projection dimension derived in \citet{lin2020convergences}, the required projection dimension for the optimal rates remains the same as in \citet{lin2020convergences} for randomized sketches and ALS Nystr\"{o}m, while the source range extends from $s\geq 1-1/\beta$ to $s> \alpha_0-1/\beta$. For plain Nystr\"{o}m, our results reduce the required projection dimension  from $m\gtrsim n^{\beta/(s\beta+1)}$ when  $s\in [1-1/\beta,2)$ to $m\gtrsim n^{\alpha\beta/(s\beta+1)}$ when $s\in (\alpha_0-1/\beta,2)$, from $m\gtrsim n^{\frac{\beta(s-\gamma)}{(2-\gamma)(s\beta+1)}} $ to $m\gtrsim n^{\frac{\alpha\beta(s-\gamma)}{(2-\gamma)(s\beta+1)}}$ when $s\geq 2$. In the most benign case $\alpha_0=1/\beta$, the required projection dimension further reduces to those in randomized sketches and ALS Nystr\"{o}m. These improvements stem primarily from our refined analysis of operator similarity and sample variance control.
\end{remark}

\begin{remark}\label{rem2}
For Tikhonov regularization with $\tau=1$, we note that when $s>2$, the saturation effect appears, where the learning rates do not depend on the source index $s$ anymore and coincide with those obtained for $s=2$, i.e., $n^{-\frac{2\beta}{2\beta+1}}$ in $L_{\rho_{\cal X}}^2$ norm. This phenomenon is well known in the theory of inverse problems \citep{bauer2007regularization} and recently has been proved for KRR \citep{lisaturation}. By exploiting the smoothness of the regression function beyond
$s=2$, \citet{lu2024saturation} and \citet{li2024generalization} further showed that, for spectral
algorithms with finite qualification $\tau$, the learning rates in the
$L_{\rho_{\mathcal X}}^{2}$ norm are $n^{-\frac{2\tau\beta}{2\tau\beta+1}}$ whenever
$s>2\tau$. To the best of our knowledge, there are currently no results on
Sobolev norm learning rates for spectral algorithms in the regime $s>2\tau$.
We conjecture that a similar saturation effect occurs when $s>2\tau$, but we
leave this question for future work. In this paper, we therefore restrict our
analysis to $s\in(0,2\tau]$.
\end{remark}

 \section{Numerical Experiments}\label{app:numerical}
In this section, we provide some simulated examples to validate our theoretical findings and real-world applications to demonstrate the effectiveness of SARP.

\subsection{Simulated Examples}
Our aim is to verify the optimal rates of SARP derived in Corollary \ref{cor1}. We consider three types of spectral algorithms: 
\begin{itemize}
\item [(1)] Kernel Ridge Regression (KRR) with $g_{\lambda}(t)=(t+\lambda)^{-1}$ and $\tau=1$; 
\item [(2)] Gradient Method (GM) with $g_{\lambda}(t) = \sum_{k=1}^{l}(1-t)^{l-k} $, $\lambda=\frac{1}{l}$ and $\tau$ being any positive number; 
\item  [(3)] Spectral Cut-off (SC) with $g_{\lambda}(t)=t^{-1}\mathbb{I}_{[t\geq \lambda]}$ and $\tau$ being any positive number.
\end{itemize}
For the simulation setup, we suppose the input space ${\cal X}=[0,1]$ and the marginal distribution $\rho_{\cal X}$ is uniform on ${\cal X}$. The kernel function is chosen as 
$
K(x,x^{\prime})=\min\{x,x^{\prime}\}.
$
The corresponding RKHS is the first order Sobolev space ${\cal H}^{1}([0,1])$ with eigenfunction-eigenvalue pairs given by
\begin{align*}
e_k(x) = \sqrt{2}\sin \frac{(2k-1)\pi x}{2},  \mu_k = \left(\frac{2}{(2k-1)\pi}\right)^2,   \forall  k\geq 1.
\end{align*}
It is easy to verify that Assumption \ref{ass2} holds with $\beta=2$, and Section III of the Supplementary Material has shown that the RKHS is benign with $\alpha_0=1/\beta=0.5$ since the eigenfunctions are uniformly bounded. The regression function is constructed as
\begin{align*}
f_{\rho}(x)=\sum_{k=1}^{\infty}\frac{1}{k^{s+0.5}\log (k+1)}e_{2k-1}(x) \quad \hbox{for some $s>0$.}
\end{align*}
We can also verify that $f_{\rho}\in [{\cal H}_K]^s_{\rho}$.
The output $y$ is generated by the following model
\begin{align*}
y=f_{\rho}(x)+\varepsilon,
\end{align*}
where $\varepsilon$ follows the standard Gaussian distribution.

We investigate three distinct scenarios: a mis-specified case with $s=0.5$ and two well-specified cases with $s=1.5$ and $s=2$. For each scenario, we vary the training sample size $n$ across 10 values evenly from $\{1000,2000,\ldots,10000\}$. The regularization parameter is set according to the theoretical optimal form $\lambda = C_{\lambda} n^{-\frac{\beta}{s\beta+1}}$. For each generated dataset, we perform a grid search of 10 values for $C_{\lambda}$  from ${10}^{-7}$ to ${10}^{2}$. We select the value of $C_\lambda$ that gives the smallest error on an independent validation set and report the corresponding performance. For the random projection methods, we use Gaussian sketching (GAU), ROS sketching (ROS), Nystr\"{o}m with uniform sampling (Ny-plain), and Nystr\"{o}m with approximate leverage scores (Ny-ALS) \citep{rudi2018fast}. For each scenario,  the projection dimension $m$ is chosen based on the theoretical requirements outlined in Corollary \ref{cor1}. \change{Here, we choose the projection dimension $m$ according to Corollary \ref{cor1} with the multiplicative constant fixed at $C_m=1$, and round the resulting value up to the nearest integer. Since our experiments mainly aim to verify the theoretical learning rates, we do not further tune $C_m$ to avoid additional computational costs.} \change{We further consider the full versions of KRR, GM, and SC as baseline methods. For a fair comparison, the full and projected estimators use the same spectral filters, kernel functions, data splits, and regularization parameter selection procedures.} We further investigate the effect of projection dimension $m$ on the performance of SARP in the real applications in the next section.

\begin{figure}
    \centering
\includegraphics[width=\textwidth]{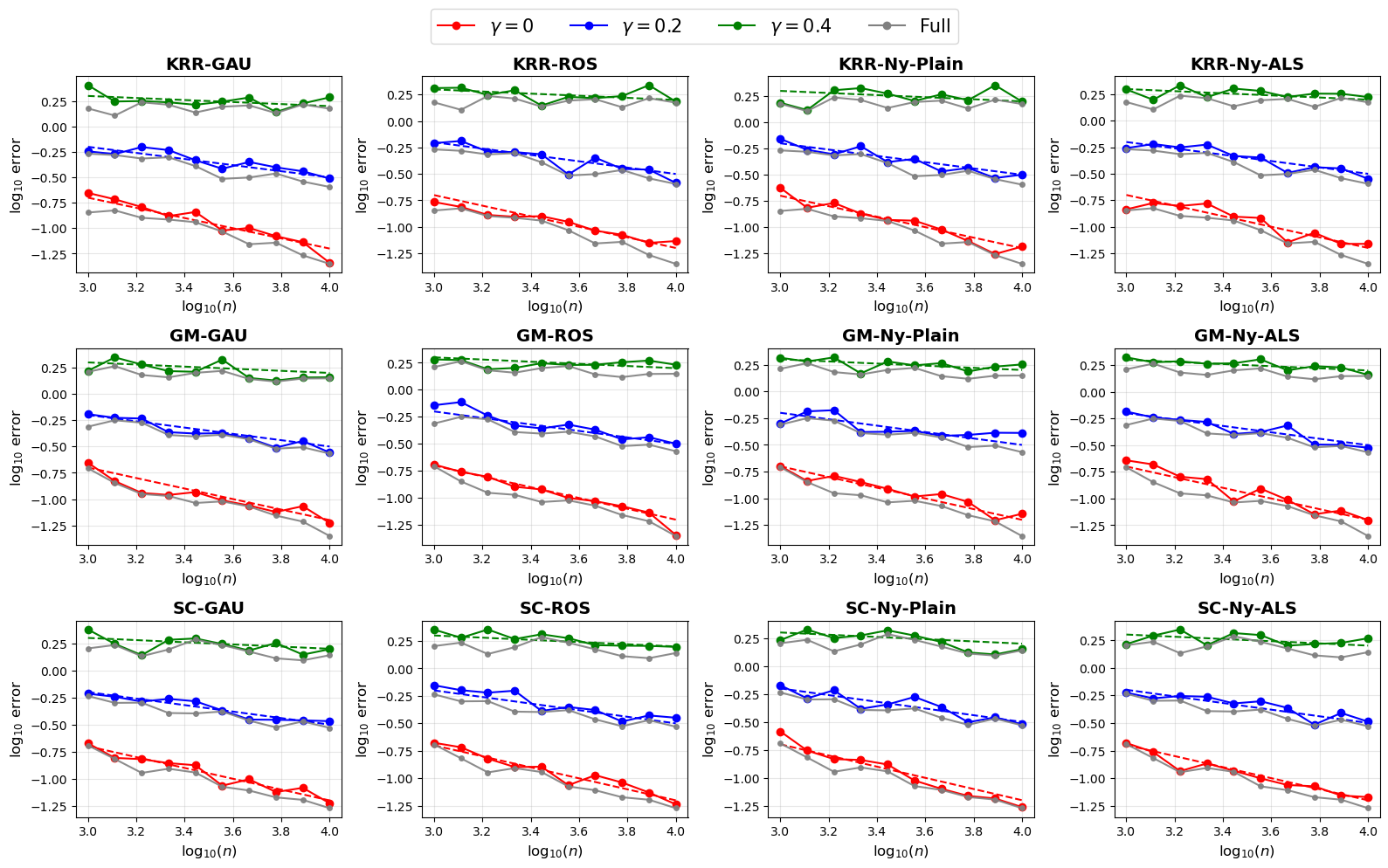} 
\vspace{-5mm}
    \caption{Log empirical learning rates of SARP with different spectral algorithms and projection methods for the mis-specified case when $s=0.5$ {averaged over 50 independent trials.} The dashed lines represent the theoretical slopes. \change{The gray lines represent the full spectral algorithms.}}
    \label{err_1}
\end{figure}

\begin{figure} 
    \centering
\includegraphics[width=\textwidth]{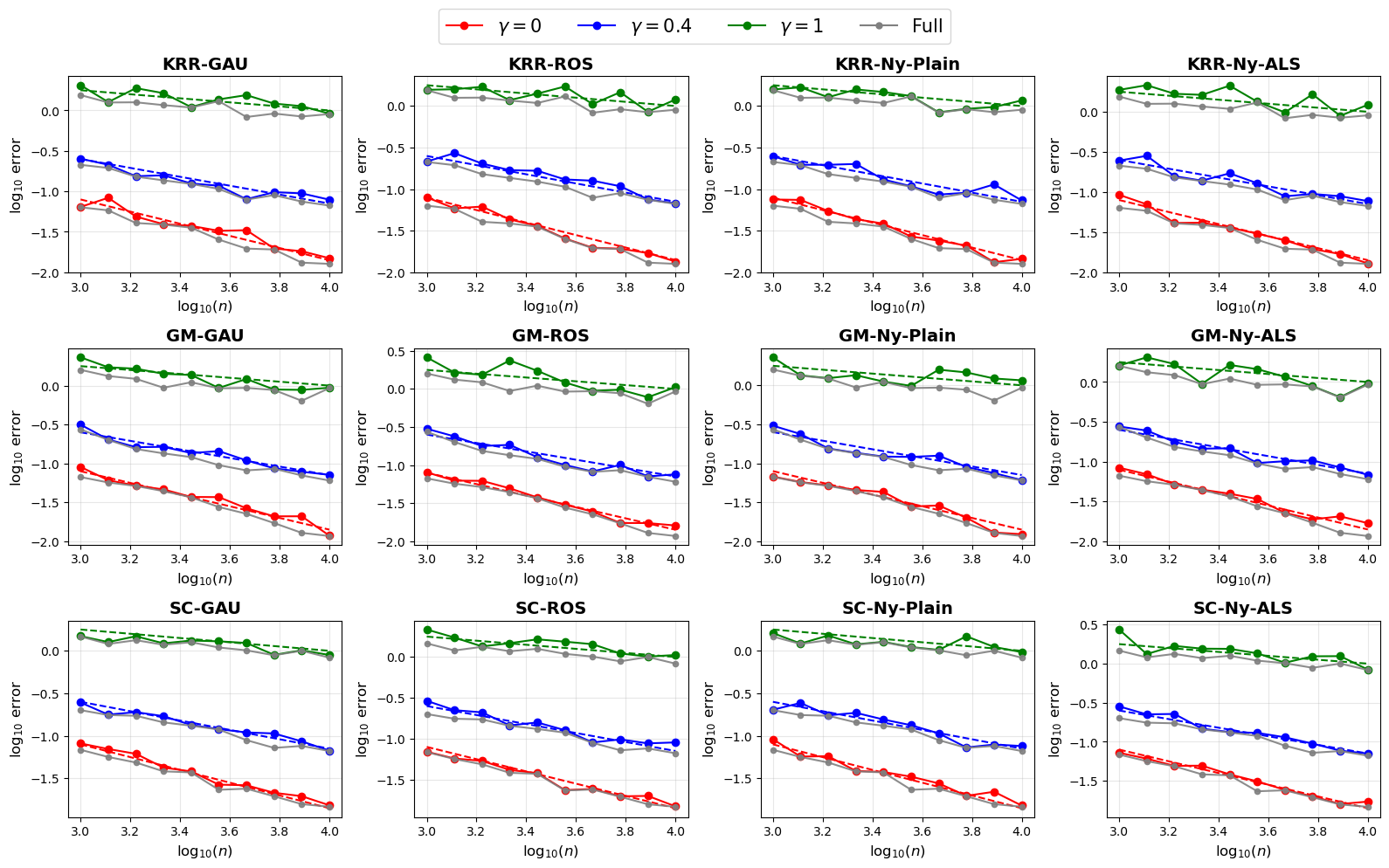} 
\vspace{-5mm}
    \caption{Log empirical learning rates of SARP with different spectral algorithms and projection methods for the well-specified case when $s=1.5$ {averaged over 50 independent trials.} The dashed lines represent the theoretical slopes. \change{The gray lines represent the full spectral algorithms.}}
    \label{err_2}
\end{figure}

\begin{figure} 
    \centering
\includegraphics[width=\textwidth]{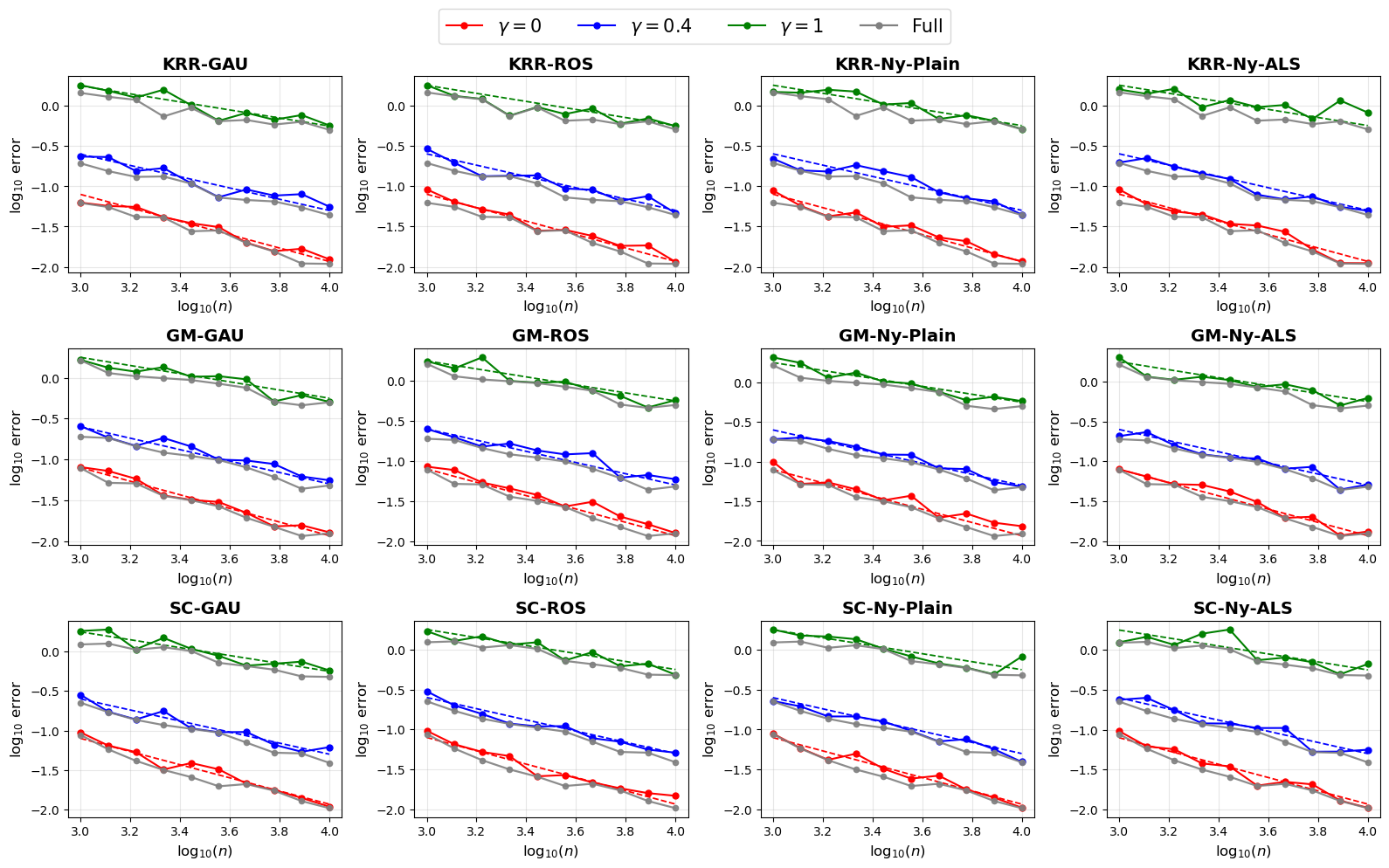} 
\vspace{-5mm}
    \caption{Log empirical learning rates of SARP with different spectral algorithms and projection methods for the well-specified case when $s=2$ {averaged over 50 independent trials.} The dashed lines represent the theoretical slopes. \change{The gray lines represent the full spectral algorithms.}}
    \label{err_3}
\end{figure}

Suppose the solution of SARP is $f^{rp}_{n,\lambda} = \sum_{i=1}^n a_{i} K(\bfx_i,\cdot)$, then learning rates $\|f_{n,\lambda}^{rp}-f_{\rho}\|^2_{[{\cal H}]_{\rho}^{\gamma}}$ can be computed as
\change{\begin{align*}
&\|f_{n,\lambda}^{rp}-f_{\rho}\|^2_{[{\cal H}]_{\rho}^{\gamma}}
=\|L_K^{-\frac{\gamma}{2}}(f_{n,\lambda}^{rp}-f_{\rho})\|_{\rho}^2=\left\|L_K^{-\frac{\gamma}{2}}\left(\sum_{i=1}^n a_i\sum_{j=1}^{\infty}\mu_j e_j(\bfx_i)e_j-\sum_{k=1}^{\infty}\frac{1}{k^{s+0.5}\log (k+1)}e_{2k-1}\right)\right\|_{\rho}^2\\
&=\sum_{k=1}^{\infty}\left(\sum_{i=1}^n a_i e_{2k-1}(\bfx_i)-\frac{1}{\mu_{2k-1}k^{s+0.5}\log (k+1)}\right)^2\mu_{2k-1}^{2-\gamma}
+\sum_{k=1}^{\infty}\left(\sum_{i=1}^n a_i e_{2k}(\bfx_i)\right)^2\mu_{2k}^{2-\gamma}.
\end{align*}}
In this paper, we use the following truncation to approximate the learning rate
\begin{align}\label{app_err}
 \change{\|f_{n,\lambda}^{rp}-f_{\rho}\|^2_{[{\cal H}]_{\rho}^{\gamma}} \approx \sum_{k=1}^{T}\left(\sum_{i=1}^na_ie_{2k-1}(\bfx_i)-\frac{1}{\mu_{2k-1} k^{s+0.5}\log (k+1)}\right)^2\mu_{2k-1}^{2-\gamma}+\sum_{k=1}^{T}\left(\sum_{i=1}^n a_i e_{2k}(\bfx_i)\right)^2\mu_{2k}^{2-\gamma}.}
\end{align}
Here we set $T=10000$. For the mis-specified case where $s=0.5$, we evaluate the Sobolev learning rates using $\gamma \in \{0,0.2,0.4\}$; for the well-specified case ($s=1.5$ and $s=2$), we use $\gamma \in \{0,0.4,1\}$.  To empirically validate the learning rates derived in Corollary \ref{cor1}, we plot the test error against the training sample size $n$ on a log-log scale. \change{The theoretical learning rate has a slope of $-\frac{(s-\gamma)\beta}{s\beta+1}$ on the log-log scale. Accordingly, we plot dashed reference lines with the theoretical slopes fixed and their intercepts determined by least-squares fitting to the log-transformed empirical results.} All the results are averaged over 50 independent trials.

As shown in Figures \ref{err_1}, \ref{err_2}, and \ref{err_3}, the empirical learning rates of SARP align well with the theoretical findings in Corollary \ref{cor1} for both mis-specified and well-specified cases. Since the theoretical learning rates are independent of the choice of spectral algorithms and projection methods once the projection dimension is chosen optimally, the empirical results also exhibit this consistency. For different Sobolev norms, we can see that the learning curves flatten as $\gamma$ increases, and the $L^2_{\rho_{\cal X}}$ error with $\gamma=0$ converges the fastest, which is consistent with our theoretical findings. \change{We also observe that SARP exhibits learning rates comparable to those of the corresponding full spectral algorithms across different projection methods and Sobolev norms.}

We note that the regression function $f_{\rho}$ is not uniformly bounded when $s=0.5$, since $e_{2k-1}(1) \equiv  \sqrt{2}$. This example confirms that the assumption of uniform boundedness on $f_{\rho}$ 
is not necessary for deriving the optimal learning rates of SARP.

\subsection{Real Applications}
In this section, we apply SARP to two real-world datasets from the UCI repository: the Handwritten Digits USPS dataset (\url{https://www.kaggle.com/datasets/bistaumanga/usps-dataset/data})
 and the cpusmall dataset (\url{https://www.csie.ntu.edu.tw/~cjlin/libsvmtools/datasets/regression.html})
. The USPS dataset contains $7291$ training and $2007$ test images of size $16 \times 16$ pixels, while the cpusmall dataset consists of $8192$ samples with $12$ CPU characteristic features, where the target variable represents the percentage of time that CPUs run in user mode. For both datasets, we randomly select $80\%$ of the samples for training and the remaining $20\%$ for testing. Note that the USPS dataset corresponds to a classification task, whereas the cpusmall dataset corresponds to a regression task.

We implement SARP using a Gaussian RBF kernel with bandwidth $24$, combined with KRR, GM, and SC as the spectral algorithms, and GAU, ROS, Ny-plain, and Ny-ALS as the projection methods. The regularization parameter $\lambda$ is chosen from $\{10^{-6},10^{-5},\dots,10^{0}\}$ via the NNI toolbox (\url{https://github.com/microsoft/nni}). The projection dimension $m$ takes values in $\{50, 100, 200, 400, 600, 800, 1000\}$. For each pair $(\lambda, m)$, we run SARP independently $20$ times and report the average performance. \change{The full SA implementations without projection are included as baselines.} Classification error rate is used as the evaluation metric for the USPS dataset, while mean squared error (MSE) is used for the cpusmall dataset.

\begin{figure} 
    \centering
\includegraphics[width=\textwidth]{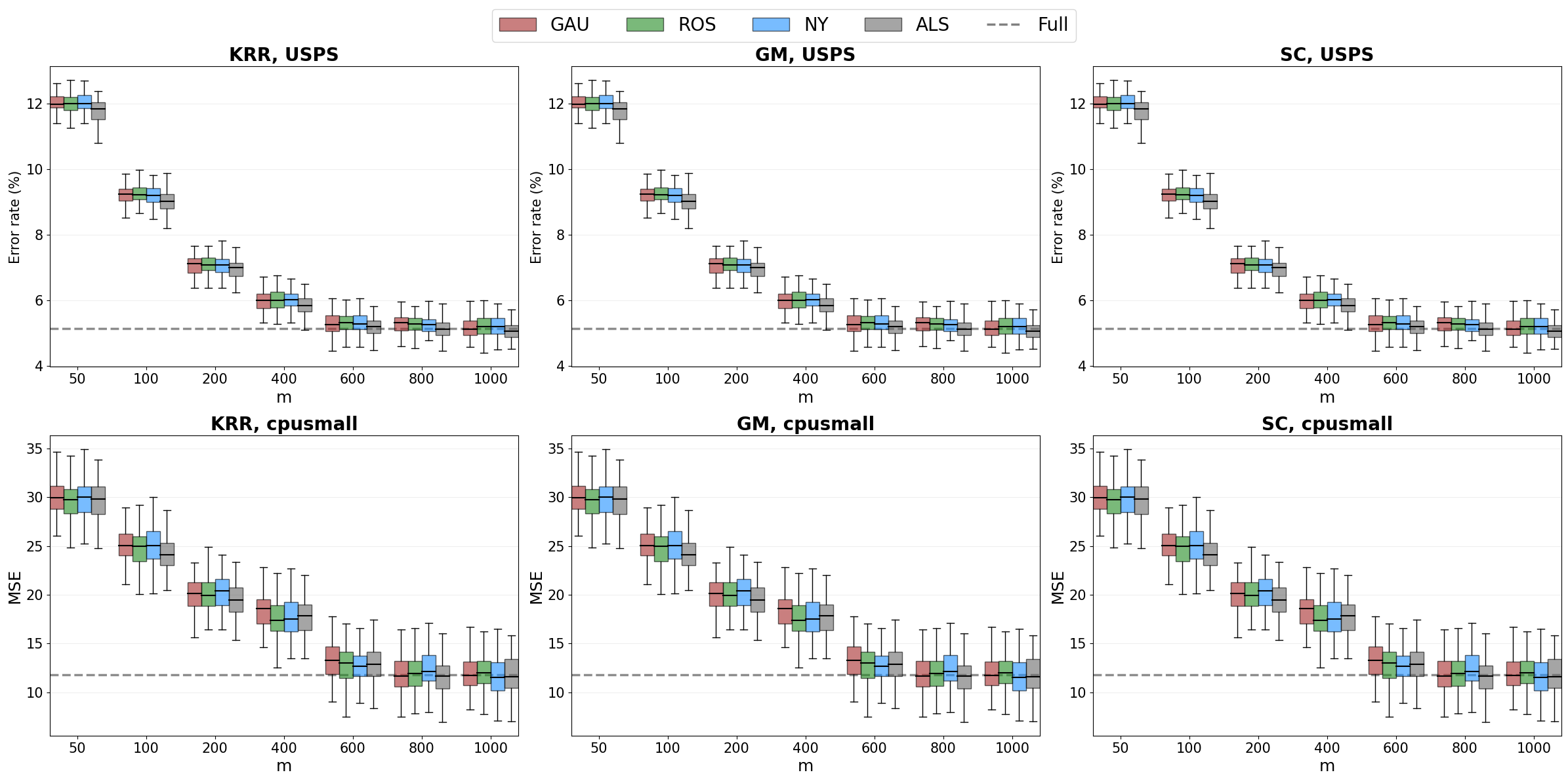} 
\vspace{-5mm}
    \caption{Performance of SARP with different spectral algorithms and projection methods on the USPS and cpusmall datasets. \change{The gray lines represent the full spectral algorithms.}}
    \label{err_4}
\end{figure}

Figure \ref{err_4} illustrates the performance of SARP with different spectral algorithms and projection methods on the USPS and cpusmall datasets. The results show that as the projection dimension $m$ increases, SARP’s performance improves across all combinations of spectral algorithms and projection methods. Notably, when $m$ reaches $600$, SARP achieves stable performance, demonstrating its ability to reduce computational costs while maintaining high accuracy. Among the various projection methods, the sketching methods (GAU and ROS) and Nyström methods (Ny-plain and Ny-ALS) exhibit comparable performance, with no significant difference between them. However, Ny-ALS slightly outperforms Ny-plain, which is consistent with our theoretical findings that Ny-ALS requires a smaller projection dimension than Ny-plain to achieve optimal learning rates. \change{As the projection dimension $m$ increases, the classification error rates and MSEs of SARP gradually approach those of the corresponding full spectral algorithms. This is expected, since a larger projection dimension generally provides a more accurate approximation of the full kernel matrix, thereby reducing the approximation error introduced by random projection.}

\change{To evaluate the computational efficiency of different projection methods against the full SA algorithm without projection, we conduct a benchmark using the USPS dataset with projection dimension $m=500$.  All methods use a Gaussian RBF kernel with bandwidth $24$ and a regularization parameter of $\lambda=10^{-2}$. Each configuration is executed independently 20 times, and the average computational time and standard deviation are reported in Table \ref{tab:computational_time}. The results show that Ny-plain and Ny-ALS generally achieve lower computational costs than GAU and ROS. The SARP methods are substantially more efficient than their full SA counterparts, particularly for KRR and SC. However, for GM, GAU and ROS require more computational time than the full SA method. This may be attributed to the relatively low computational cost of the iterative GM algorithm, where the additional cost of constructing and applying random projections can outweigh their computational benefits. In contrast, Ny-plain and Ny-ALS remain more efficient than the full-kernel GM algorithm. Overall, these results suggest that suitable projection methods can significantly accelerate computation while maintaining comparable predictive performance.}

\begin{table}[htbp]
\centering
\caption{\change{Computational time (in seconds) of different
spectral algorithms and projection methods
for $m=500$.}}
\label{tab:computational_time}

\small
\setlength{\tabcolsep}{5pt}
\renewcommand{\arraystretch}{1.1}

\begin{tabular}{cccccc}
\toprule
Algorithm & GAU & ROS & NY & ALS & Full \\
\midrule
KRR
& $0.825 \pm 0.163$
& $1.339 \pm 0.612$
& $0.258 \pm 0.076$
& $0.238 \pm 0.010$
& $2.425 \pm 0.237$ \\

GM
& $0.883 \pm 0.198$
& $1.332 \pm 0.628$
& $0.230 \pm 0.027$
& $0.274 \pm 0.002$
& $0.661 \pm 0.012$ \\

SC
& $0.899 \pm 0.147$
& $1.391 \pm 0.590$
& $0.309 \pm 0.030$
& $0.267 \pm 0.011$
& $3.342 \pm 0.103$ \\
\bottomrule
\end{tabular}
\end{table}

 \section{Conclusion}\label{sec:conclusion}
In this paper, we have established high-probability Sobolev norm error bounds for SARP under general source conditions. The results demonstrate that, by selecting an appropriate projection dimension and regularization parameter, these algorithms can achieve minimax optimal learning rates up to a logarithmic factor. Notably, our analysis extends to the mis-specified case with $s> \alpha_0-1/\beta$ and Sobolev norms $\gamma \in [0,1 \wedge s]$, which were not fully addressed in previous works. Our findings contribute to a deeper understanding of the theoretical properties of SARP and provide practical insights for its implementation in large-scale learning tasks. Future research may proceed in several directions. 
First, our analysis techniques may apply to other scalable kernel learning algorithms, such as random features \citep{rudi2017generalization,wang2024optimal} and stochastic gradient methods \citep{lin2016optimal}. Second, it is natural to extend our analysis to vector-valued settings, for example, in connection with conditional mean embeddings \citep{li2022optimal}. Third, one may relax the noise assumption to the heavy-tailed case and combine our SARP framework with robust concentration tools, which have recently been developed by \citet{mollenhauer2025regularized}. Finally, a fully general information-theoretic lower bound on the projection dimension $m$ for SARP remains an open problem.




 \bibliography{sample}
\bibliographystyle{cas-model2-names}

\end{document}